\documentclass[10pt,twocolumn,letterpaper]{article} 

    \usepackage{geometry}
    \usepackage{cvpr}
    \usepackage{times}
    \usepackage{epsfig}
    \usepackage{graphicx} 
    \usepackage{amsmath}
    \usepackage{amssymb} 
 
    \usepackage{mathrsfs}
    \usepackage{enumitem}
    \usepackage[linesnumbered,ruled,vlined]{algorithm2e}%[ruled,vlined]{  
    \usepackage{algpseudocode}  
    \usepackage{amsmath}
 
    \usepackage{epsfig}
    \usepackage{bm}
    \usepackage{multirow} %multirow for format of table  
    \usepackage{booktabs}
    \usepackage[noblocks]{authblk}
    \usepackage{array}
    \usepackage{subfigure}
    \usepackage{epstopdf}
    \usepackage{color}
    \usepackage{epstopdf}
    \usepackage[normalem]{ulem}  
    \usepackage{float} 
    \usepackage{lipsum}
    \newcommand{\erhao}{\fontsize{21pt}{\baselineskip}\selectfont}

    \usepackage[breaklinks=true,bookmarks=false]{hyperref}

    \cvprfinalcopy 
    
    \def\cvprPaperID{2693} % *** Enter the CVPR Paper ID here

    \ifcvprfinal\fi
\begin{document}

    {\onecolumn

    \noindent \textbf{\erhao{Weakly Supervised Open-set Domain Adaptation \\ by Dual-domain Collaboration}}

    \vspace{2cm}

    \noindent {\LARGE{Shuhan Tan, Jiening Jiao, Wei-Shi Zheng}}

    %\Large
    \vspace{2cm}

    \vspace{1cm}

    \noindent For reference of this work, please cite:

    \vspace{1cm}
    \noindent Shuhan Tan, Jiening Jiao, Wei-Shi Zheng.
    ``Weakly Supervised Open-set Domain Adaptation by Dual-domain Collaboration''
    In \emph{Proceedings of the IEEE International Conference on Computer Vision and Pattern Recognition (CVPR).} 2019.

    \vspace{1cm}

    \noindent Bib:\\
    \noindent
    @inproceedings\{tan2019weaklysupervised,\\
    \ \ \   title=\{Weakly Supervised Open-set Domain Adaptation by Dual-domain Collaboration\},\\
    \ \ \  author=\{Tan, Shuhan and Jiao, Jiening and Zheng, Wei-Shi\},\\
    \ \ \  booktitle=\{Proceedings of the IEEE International Conference on Computer Vision and Pattern Recognition (CVPR)\},\\
    \ \ \  year=\{2019\}\\
    \}
    }

    \restoregeometry

    %%%%%%%%% TITLE
    \title{Weakly Supervised Open-set Domain Adaptation by \\ Dual-domain Collaboration}

    \author[ ]{Shuhan Tan$^{1,3,4}$}
    \author[ ]{Jiening Jiao$^{2,3}$}
    \author[ ]{Wei-Shi Zheng$^{1,3}$\thanks{Corresponding author}}

    \affil[ ]{$^1$School of Data and Computer Science, Sun Yat-sen University, China}
    \affil[ ]{$^2$School of Electronics and Information Technology, Sun Yat-sen University, China}
    \affil[ ]{$^3$Key Laboratory of Machine Intelligence and Advanced Computing, Ministry of Education, China}
    \affil[ ]{$^4$Accuvision Technology Co. Ltd.}
    \affil[ ]{\tt\small \{tanshh, jiaojn\}@mail2.sysu.edu.cn, wszheng@ieee.org}

    \maketitle
    \thispagestyle{empty}

    \newcommand\blfootnote[1]{%  
    \begingroup 
    \renewcommand\thefootnote{}\footnote{#1}% 
    \addtocounter{footnote}{-1}% 
    \endgroup 
    }

    \newcommand{\jie}[1]{{\color{blue}#1}}
    \newcommand{\shh}[1]{{\color[RGB]{0, 0, 0} #1}}
    \newcommand{\shhh}[1]{{\color{black} #1}}
    \newcommand{\rev}[1]{{\color[RGB]{0, 0, 0} #1}}

    \begin{abstract}
    In conventional domain adaptation, a critical assumption is that there exists a fully labeled \rev{domain} (source) that contains the same label space as another unlabeled or \shhh{scarcely} labeled domain (target). 
    However, in the real world, there often exist application scenarios in which both domains are partially labeled
    and not all classes are shared between these two domains. 
    Thus, it is meaningful to \rev{let} partially labeled domains \rev{learn from each other} to classify all the unlabeled samples in each domain \rev{under an open-set setting.} \rev{We} consider \rev{this problem} as \textit{weakly \shh{supervised}
        open-set domain adaptation}.
    To address  \shhh{this} practical setting,     
    we propose the \textit{Collaborative Distribution Alignment (CDA)} method, which performs knowledge transfer bilaterally and works collaboratively to classify unlabeled data and identify \rev{outlier samples}. Extensive experiments on the \textit{Office} \rev{benchmark} and \rev{an application on} 
    \rev{ person re-identification} show that our method achieves state-of-the-art performance. 
    \end{abstract}

    \section{Introduction}
    We have recently seen state-of-the-art performances in many \shhh{computer vision} applications\shhh{,} most\shhh{ly with} deep learning. Nevertheless, most of that impressive progress strongly depends on a large collection of labeled data. However, in real-world applications, labeled data are often scarce or sometimes even unobtainable. A straightforward solution is to utilize off-the-shelf labeled datasets (\textit{e.g.}, ImageNet \cite{ImageNet}). However, datasets collected in different scenarios could be in different data domains due to differences in many aspects (\textit{e.g.}, background, light and sensor resolution). Therefore, there is a strong demand to align the two domains and leverage labeled data from the source domain to train a classifier that is applicable in the target domain. This approach is termed \textit{domain adaptation} \cite{transfer_survey}. 

    \begin{figure} [t]
      \centering 
      \includegraphics[width=1.112\linewidth]{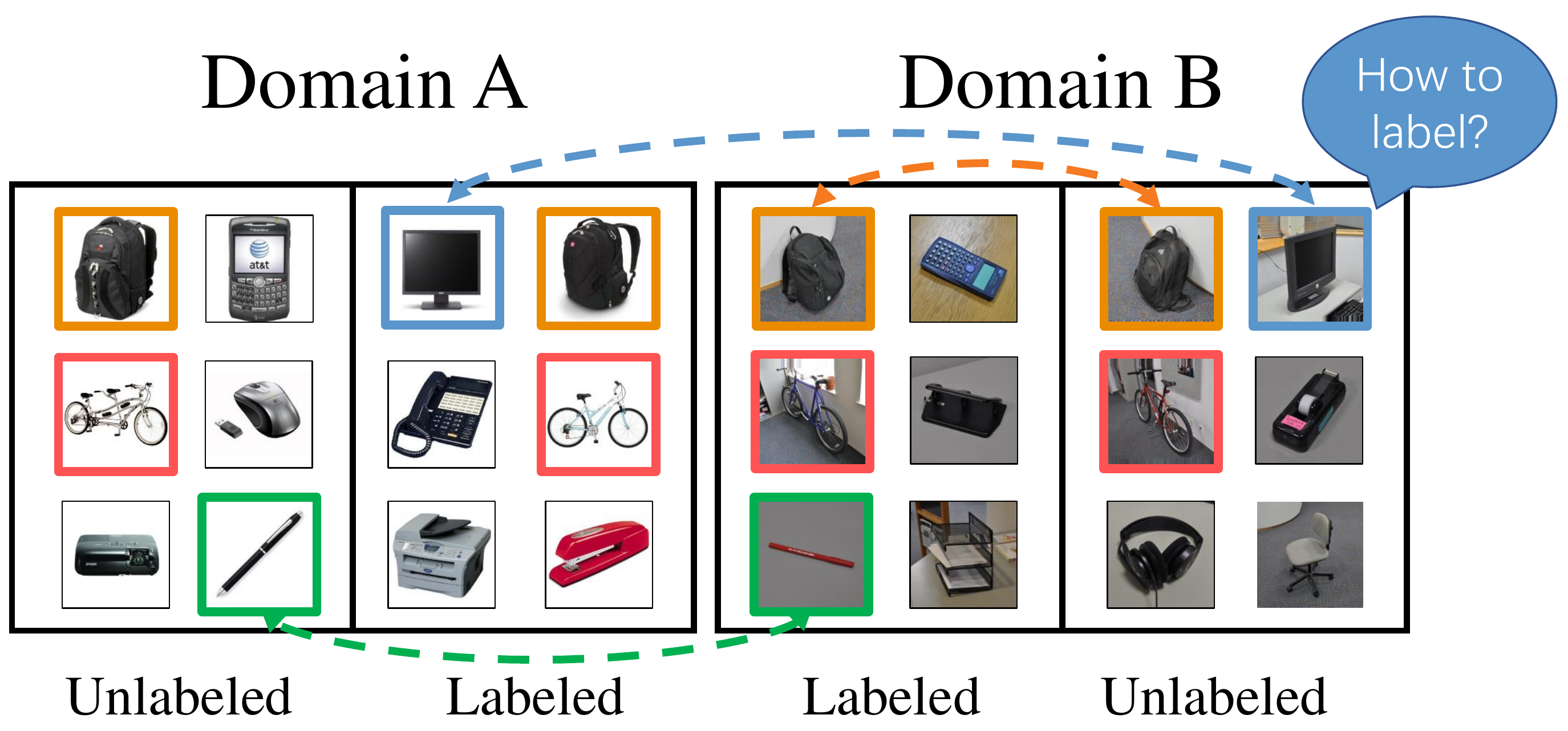}
      \caption{
        \footnotesize
            \small
        An illustration of weakly supervised open-set domain adaptation \rev{for domain collaboration}. \rev{In} each domain, there are some labeled and unlabeled samples. The frame colors represent the class of each item, while the \textit{unknown-class} items have black frames. Without domain collaboration, the blue-framed item in Domain B cannot be classified properly because there are no labeled blue-framed data in its own domain.
      This is the same for the green-framed item in Domain A. (Best viewed in color)
      }
      \label{fig:reid}
      \vspace{-0.3cm}
    \end{figure}
      \begin{figure*}[!ht]
      \centering                                      
      \subfigure[Closed-set]{
        \centering        
        \includegraphics[width=0.25\textwidth]{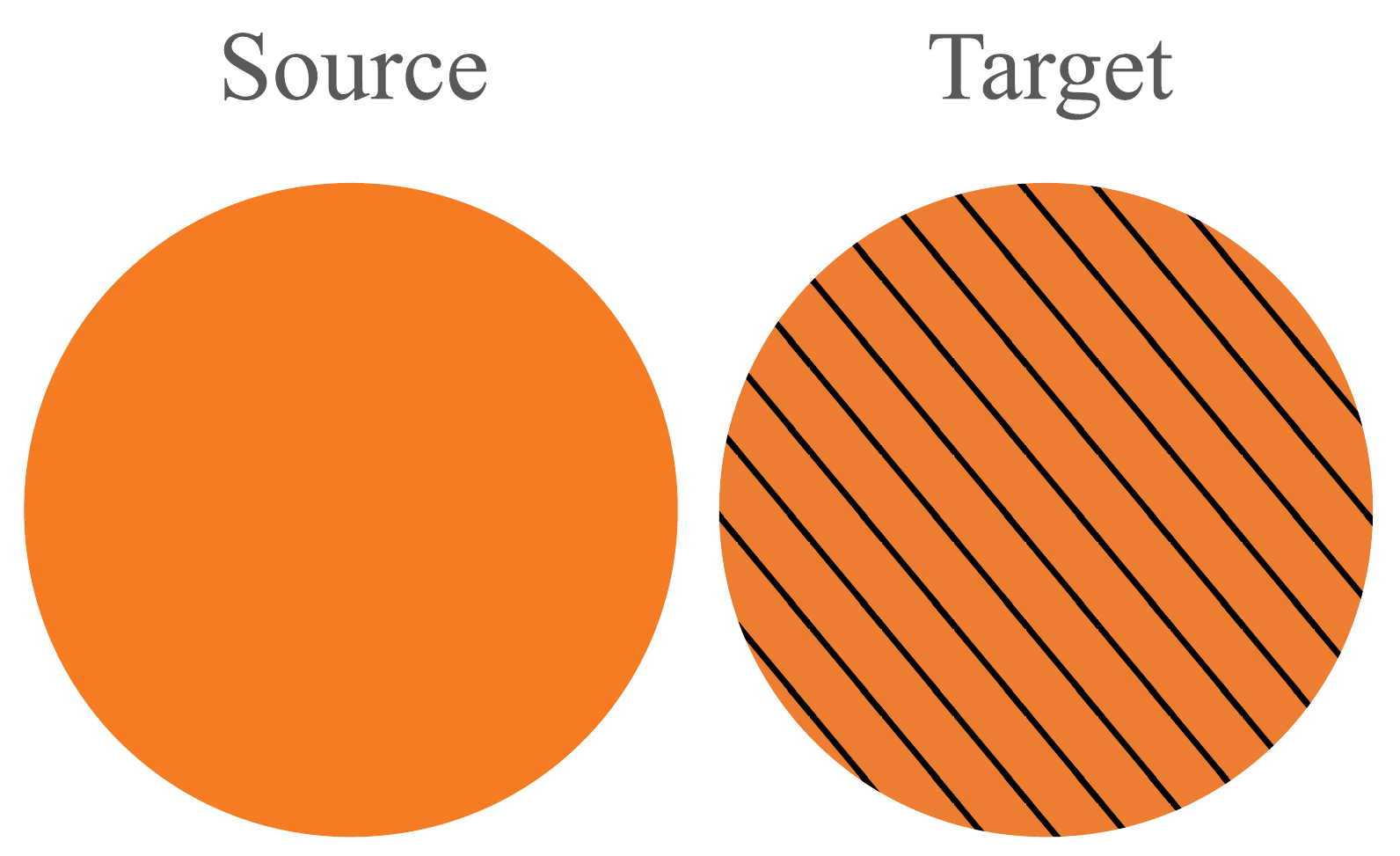}  
      }
      \hspace{0.045\textwidth}
      \subfigure[Open-set]{
        \centering        
        \includegraphics[width=0.25\textwidth]{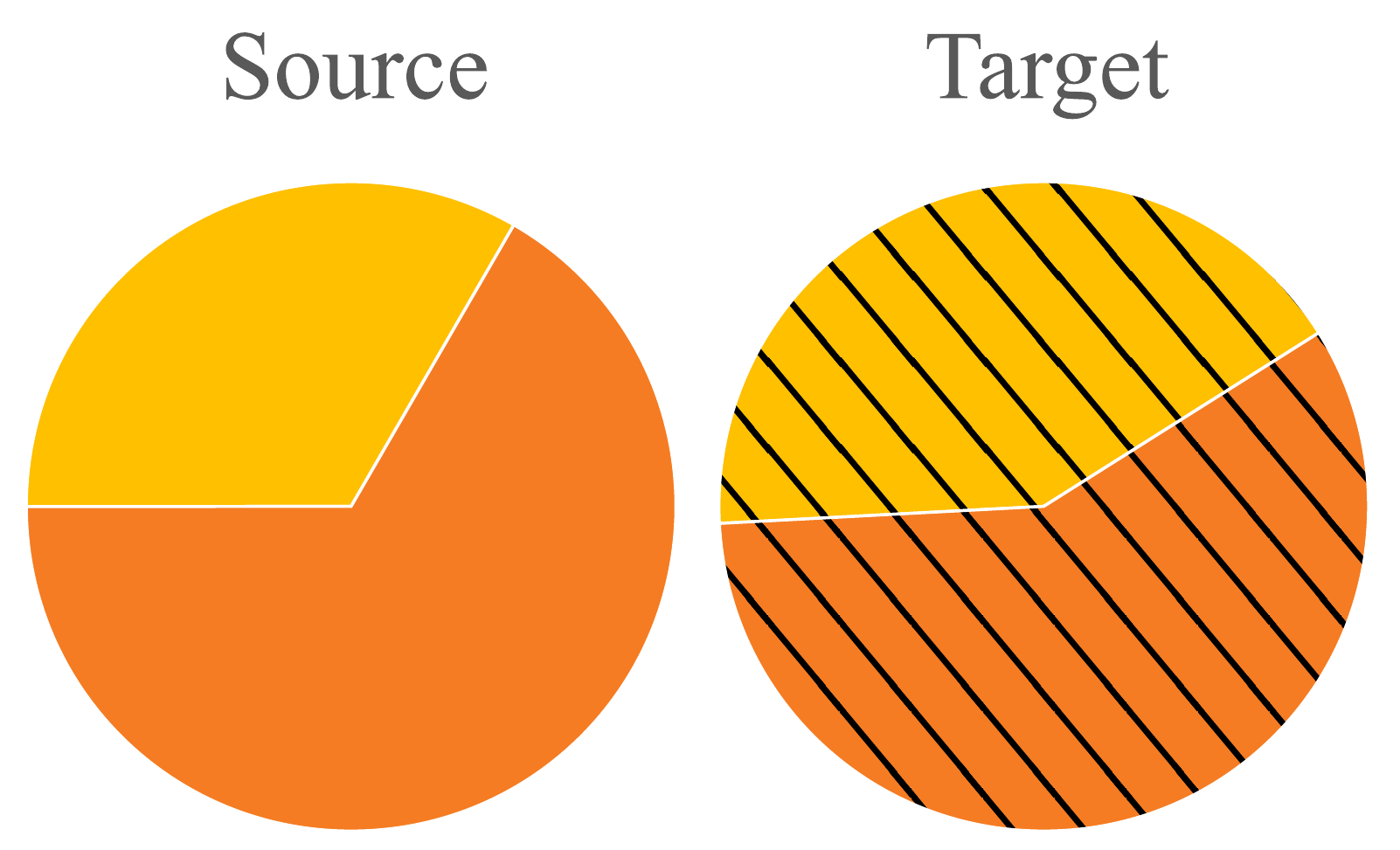}  
      }
      \hspace{0.045\textwidth} 
      \subfigure[Weakly Supervised Open-set]{
        \centering        
        \includegraphics[width=0.25\textwidth]{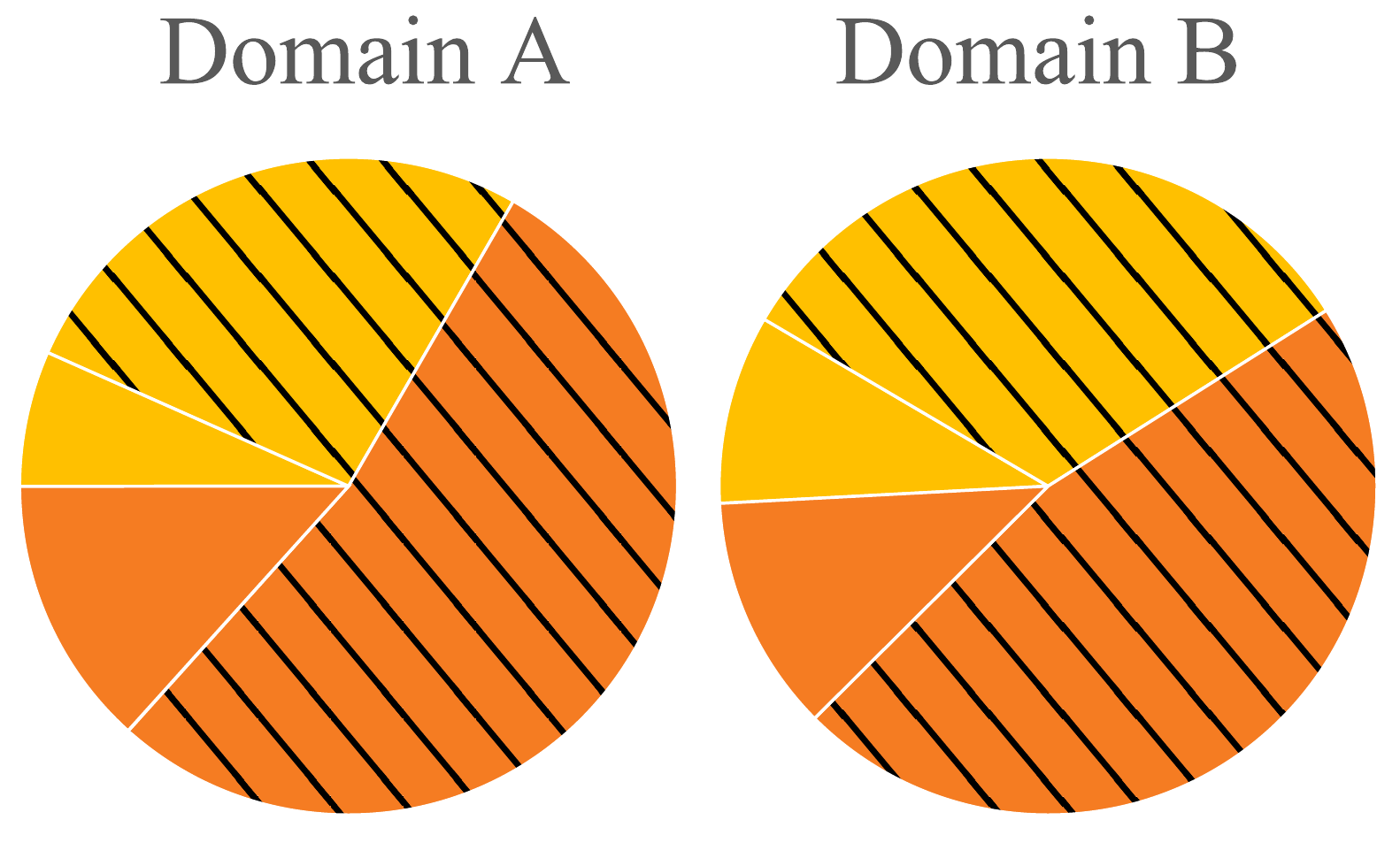}  
      }
      \caption{\textbf{(a)} closed-set domain adaptation, which assumes that the source and target domains contain the same set of \shh{interested}
      %known
      classes (orange), and the two domains are either fully labeled (pure color), or unlabeled (with \shhh{shade}). \textbf{(b)} open-set domain adaptation \cite{ATI}, which allows both domains to contain unknown-class images (yellow). \textbf{(c)} weakly supervised open-set domain adaptation, in which both domains are scarcely labeled, with unknown-class images included. (Best viewed in color)}
      \label{fig:setting}      
      \vspace{-0.5cm}
      \end{figure*}

    In conventional domain adaptation, the label spaces of the source and target domains are identical. However, such a setting may be restricted in applications. Recently in  \cite{ATI}, \textit{open-set} domain adaptation was proposed, where there exist samples in the target domain that do not belong to any class in the source domain, and vice versa. \rev{These samples cannot be represented by any class in the known label space, and they are therefore aggregated to an additional \textit{unknown class}.}
    One of our main goals is to detect these \rev{unknown-class samples} in the target domain and in the meantime still classify the remaining \rev{\textit{known-class}} samples. 

    \shhh{Moreover}, in real applications, there often exists a large number of partially labeled datasets, \shhh{where we cannot always find an off-the-shelf source dataset.} Therefore, in the absence of an ideally large source domain, it is meaningful to perform domain adaptation collaboratively between these partially labeled domains.
  Hence, we consider the open-set domain adaptation together with the partially labeled datasets, which we call \textit{weakly supervised open-set domain adaptation}. The differences between our setting and the existing ones are illustrated in Figure \ref{fig:setting}. 

  \rev{
  The weakly supervised open-set domain adaptation is practical whenever research or commercial groups need to collaborate by sharing datasets. This situation often happens when two groups each collected a large dataset in related areas, but each of them has only annotated a subset of all the item classes of interest
  (see Figure \ref {fig:reid}). To fully label the collected datasets, the groups need to share their datasets and let them learn from each other, which exactly fits our proposed setting, while traditional domain adaptation settings are not applicable. 
  
  Moreover, the proposed setting is practical when collecting a large source domain is unrealistic. This is often the case in real applications, as it is difficult to find large labeled data for items that only appear in specific tasks, which makes existing domain adaptation methods not applicable. For example, in person re-identification (see Section \ref{exp_reid}), it is impossible to find off-the-shelf labeled datasets for person identities captured by a certain camera view. Under this situation, one can let partially labeled datasets collected in different circumstances (\textit{e.g.}, data collected by different camera views) help label each other, for which our proposed setting is very suitable.}

\begin{figure*}[t]
      \centering                                      
      \subfigure[Original Domains]{
        \centering        
        \includegraphics[width=0.27\textwidth]{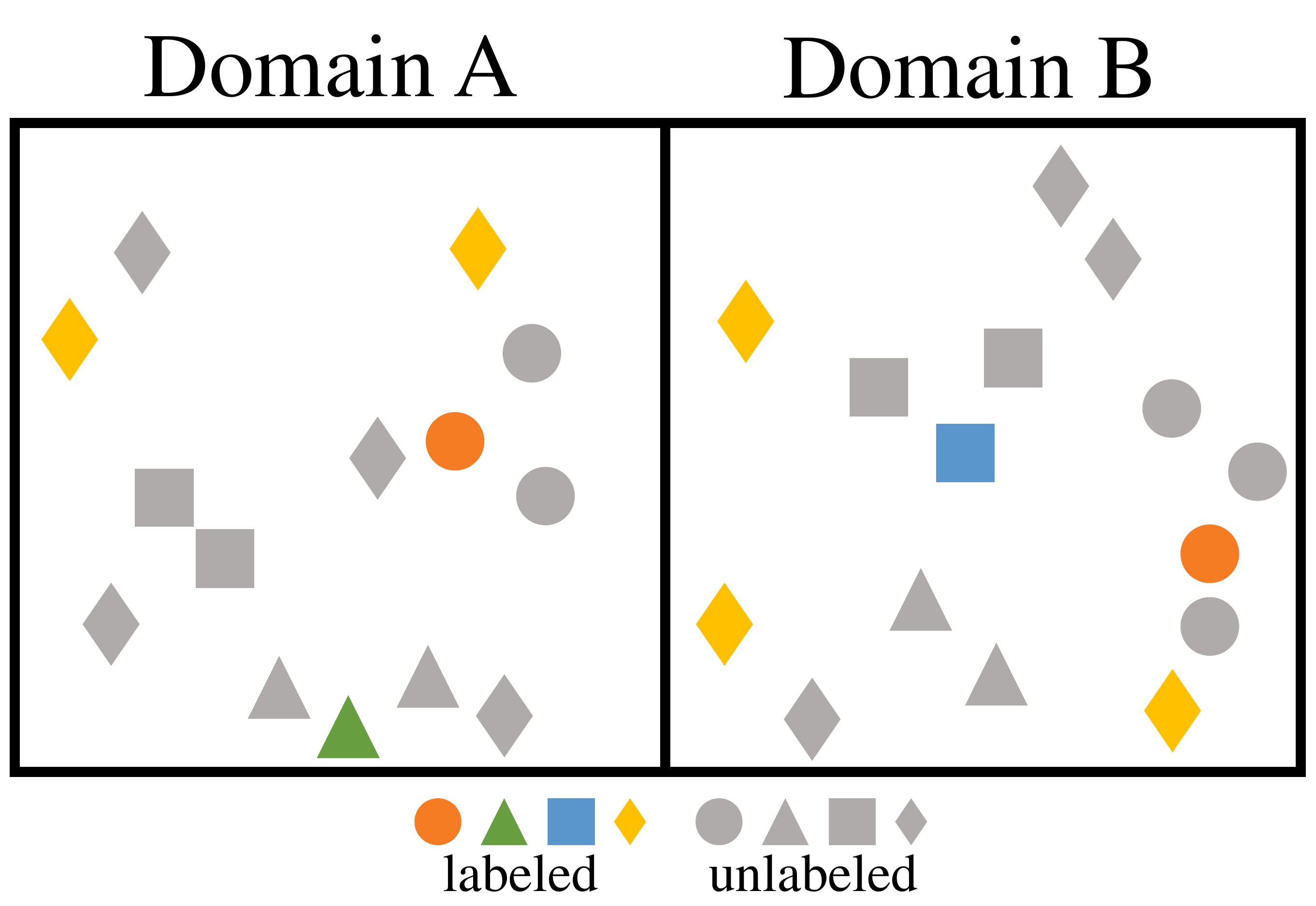}      
      }
      \subfigure[Pseudo-Label Assignment]{
        \centering        
        \includegraphics[width=0.27\textwidth]{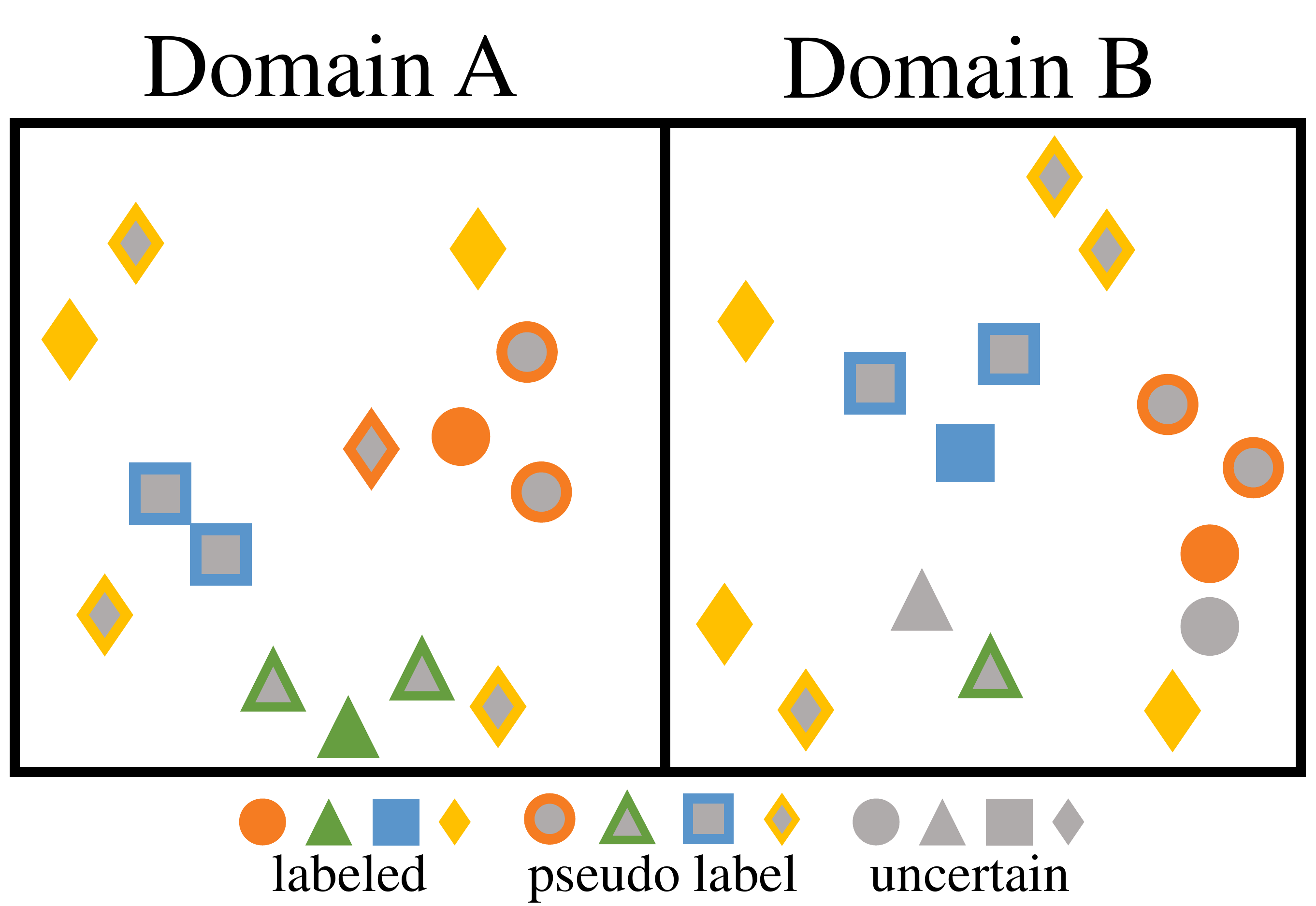}          
      }
      \subfigure[Transformed Domains]{
        \centering        
        \includegraphics[width=0.27\textwidth]{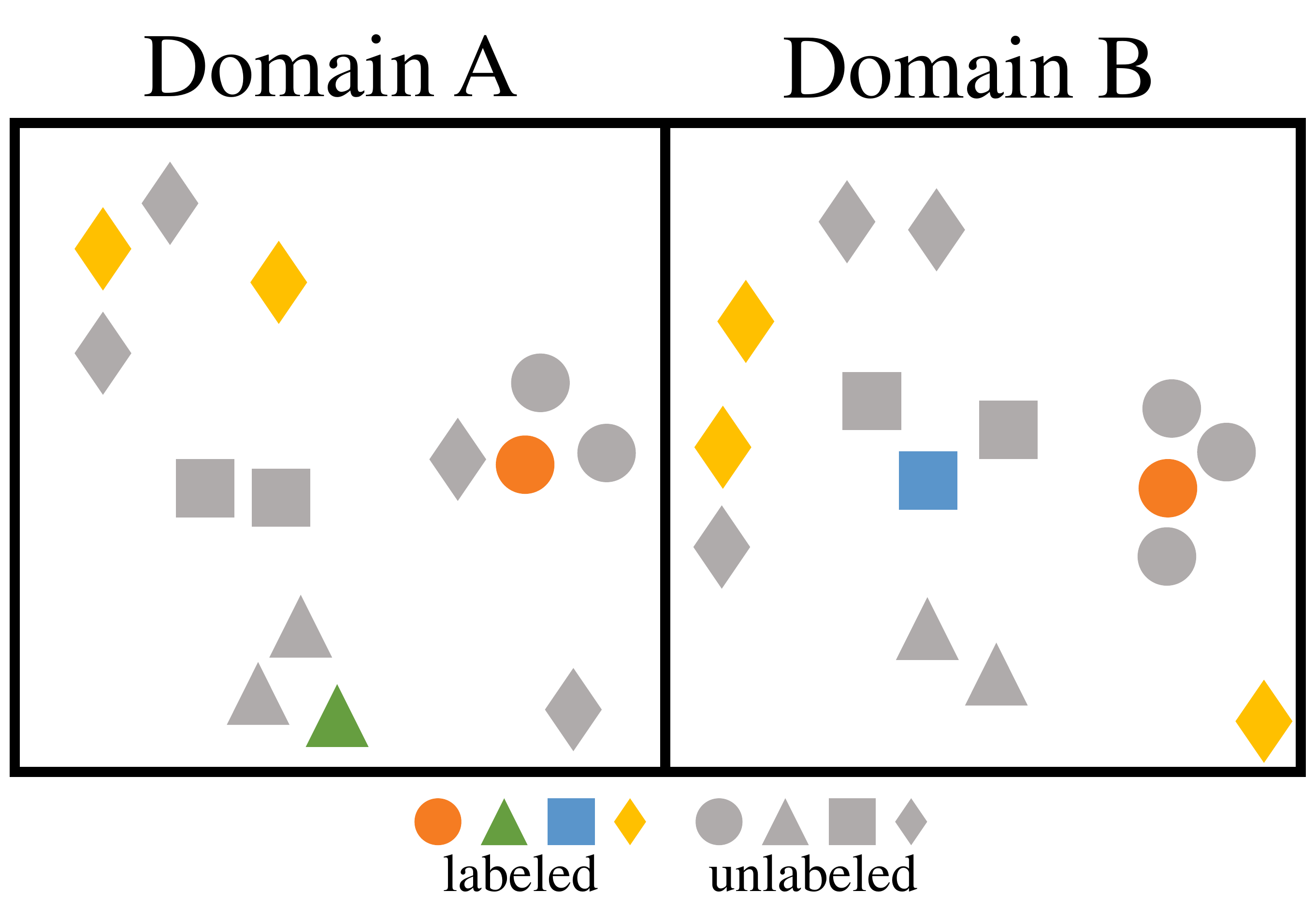}            
      } 
      \subfigure[Latent Space]{  
        \centering        
        \includegraphics[width=0.135\textwidth] {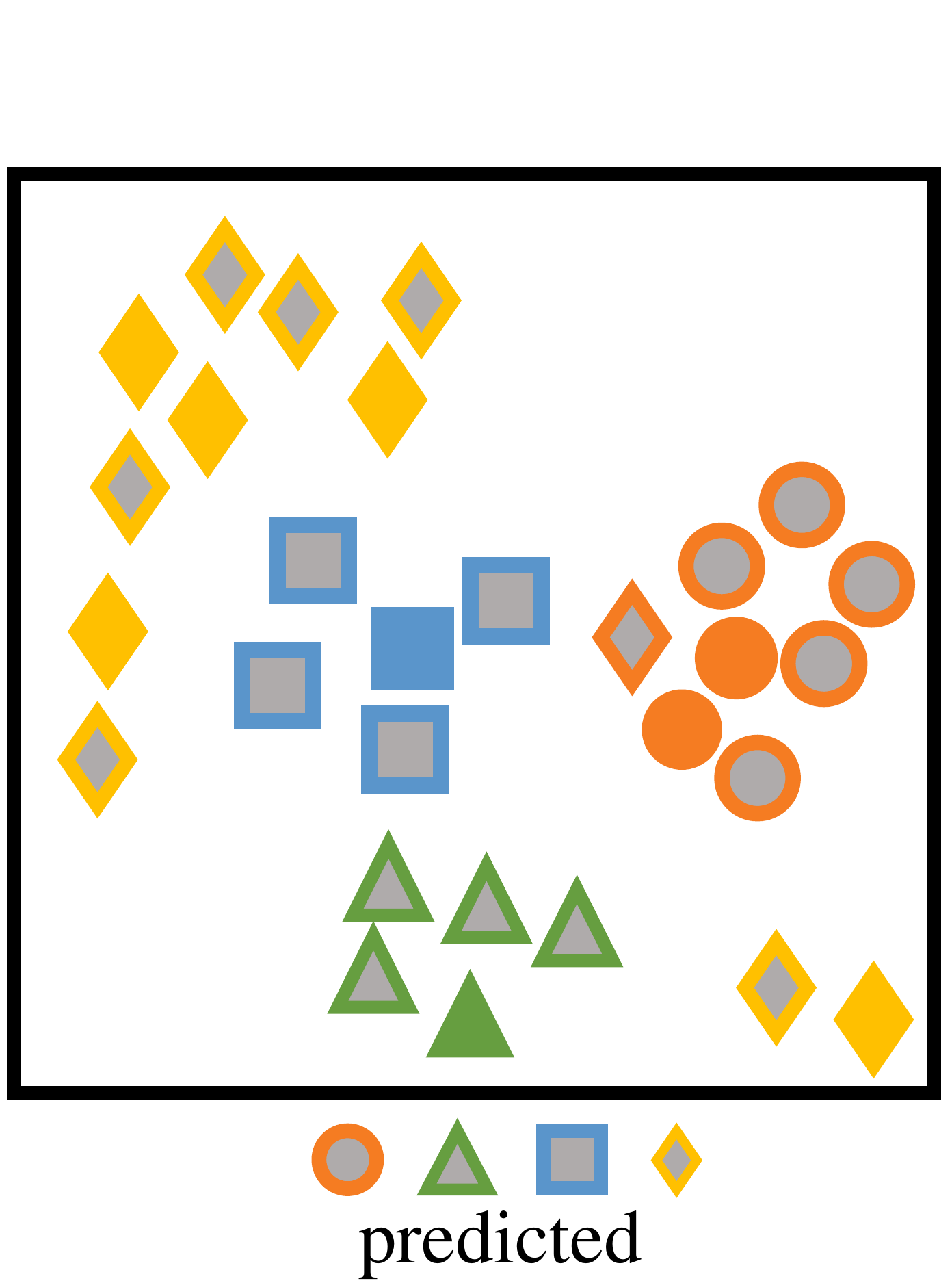}     
      }
      \caption{An overview of our approach. In \textbf{(a)}, labeled samples from different classes are filled with \rev{corresponding} colors, while the gray samples are unlabeled. The shape of each sample indicates its ground-truth class. Specifically, the diamond-shaped samples represent the \rev{unknown-class samples}. Note that the unknown-class samples are those that cannot be represented in the known label space, while the unlabeled samples are those without label information. Therefore, an unknown-class sample can be \textit{either} labeled or unlabeled.
      %  Note that the square-shaped unlabeled samples in Domain A \textit{could not} be properly labeled without \textit{collaborating} with Domain B. This also happens \shh{to}
      %is for triangles in Domain B. 
      In \textbf{(b)}, we assign \textit{pseudo-labels} (represented by frame colors) for some unlabeled samples, while excluding the remaining \textit{uncertain} samples as outliers. In \textbf{(c)}, we learn a set of domain-specific mappings that transform samples onto a latent domain, where domain discrepancy is reduced, same-class samples are aggregated, and separations between unknown and known \rev{class} samples are formed. We then use the transformed features to update pseudo-labels in (b) and iterate between (b) and (c) until convergence. Finally, in the latent space shown in \textbf{(d)}, we use a base classifier to annotate all the unlabeled samples, which 
      %needs
      \shh{we expect} to be predicted as one of the known classes (red, green, and blue), or as an unknown-class sample (yellow). (Best viewed in color)} 
      \label{fig:method}      
      \vspace{-0.5cm}
      \end{figure*}

  A key challenge of solving the weakly supervised open-set domain adaptation problem is how to perform domain adaptation when there only exists a small quantity of labeled data in both domains.
  To address this challenge, we propose to align two domains in a \textit{collaborative} fashion. By implementing a novel dual mapping method, \textit{i.e.}, learning domain-specific feature transformations for each domain, we iteratively map the feature spaces of two domains $\mathcal{D_A}$ and $\mathcal{D_B}$ onto a \shh{shared} latent space. To alleviate the effect of the unknown class, \shhh{during the dual mapping optimization,} we simultaneously maximize the margin between \textit{known-class} and \textit{unknown-class} samples\shhh{,} along with minimizing feature distribution discrepancy and intraclass variation\shhh{.} \shhh{Therefore, we are able} to obtain more robust \rev{unknown} detection under the open-set setting\shhh{.} In addition, to enrich the discriminant ability, we propose to enlarge the label set by making pseudo-label predictions for the unlabeled samples, which are further selected by information entropy \shhh{and used in dual-mapping optimization.} After the domains are aligned in the shared latent space, we finally learn a classifier with labeled data from \textit{both} domains. We call the above process that covers all the main challenges the \textit{Collaborative Distribution Alignment (CDA)} method. Figure \ref{fig:method} shows an overview of our method.

    \rev{The contributions of our paper are as follows:
    \begin{enumerate}[itemindent=0em]
    \setlength{\itemsep}{0pt}
    \setlength{\parsep}{0pt}
    \setlength{\parskip}{0pt} 
    \vspace{-0.3cm} 
    \item We proposed the weakly supervised open-set domain adaptation, which allows partially labeled domains to be fully annotated by learning from each other. This setting extends domain adaptation to the scenarios where an ideal source domain is absent.
    \item We proposed a collaborative learning method to address the proposed problem, which is well designed to handle the challenges introduced by both weakly supervised and open-set settings.
    \item We evaluated our method on the standard domain adaptation benchmark \textit{Office} and a real-world person re-identification application and showed its effectiveness. 
    \end{enumerate} }

    \vspace{-0.4cm}
    \section{Related Works}
    \noindent{\textbf{Conventional Domain Adaptation.}} Domain adaptation is an important research area in machine learning. The core task of standard domain adaptation is to transfer discriminative knowledge from one fully labeled source domain to a scarcely labeled or unlabeled target domain \cite{transfer_survey}. 
    A large number of methods have been proposed to solve this problem.
    A main stream of these works is to align the feature distributions of different domains \shh{onto}
    %in
    a latent space.
    Along these lines, TCA \cite{TCA} and JDA \cite{JDA} were proposed to explicitly align marginal or joint distributions, while several models further extended the distribution alignment with structural consistency or geometrical discrepancy \cite{hou2016unsupervised,JGSA}. 
% Quality control editor: Abbreviations and acronyms (e.g., TCA, JDA, PCA) are typically defined the first time the term is used within the main text and then used throughout the remainder of the manuscript. Please consider adhering to this convention. The target journal may have a list of abbreviations that are considered common enough that they do not need to be defined.    
    \shh{Several recent works also introduced dataset shift losses to train deep networks that learn domain-invariant features \cite{DAN, RTN, JAN, DDC}. Another stream of work focuses on learning a subspace where both domains have a shared representation.} \shhh{SA \cite{SA} was proposed to align the PCA subspace, while CORAL \cite{CORAL} utilizes second-order statistics to minimize domain shift.} \shh{Manifold learning was also conducted to find a geodesic path from the source to target domain \cite{GFK, GFS}. More recently, various generative adversarial models have been proposed \cite{Deng_2018_CVPR,Hu_2018_CVPR, OpenBP, ADDA}, in which feature extractors are trained to generate features for target samples that are not discriminated from the source samples.} Although these methods are state-of-the-art for conventional domain adaptation tasks, none of them can directly tackle the weakly supervised open-set domain adaptation problem. On one hand, many of these methods require a fully labeled domain as a source, which does not exist in our setting. On the other hand, all of them are designed for the setting where the label spaces \shh{are identical}, while in our setting, the label spaces of the two domains are overlapping but not identical. Thus, the performances of these methods are hindered by the \textit{unknown-class} samples in each domain. In contrast, our proposed method is well designed for scarcely labeled domain pairs under the open-set setting. 

  % \rev{Our proposed setting may be confused with semi-supervised domain adaptation \cite{ATI, MMDT,matasci2015semisupervised, saenko2010adapting}. The semi-supervised setting means that there exist a few labeled samples in the target domain, while a large fully-labeled source domain remains available. In contrast, the proposed setting is more challenging as both domains contain only a few labeled samples.}

    \noindent{\textbf{Open-set Setting.}} In real-word application, the unknown impostors, which cannot be represented by any class in the label space of known classes, could significantly influence the overall performance. This problem has been \shhh{receiving} attention in the field of open-set object recognition \cite{OpenFace,OpenRecognition}. A typical solution is the \shh{OSVM} \cite{OSVM}, in which impostors are detected by estimating the inclusion probability. In the context of person re-identification, Zheng \textit{et al.} proposed to explicitly learn a separation between a group of nontarget and target people \cite{OpenWorld}. 
    Recently, Busto and Gall proposed the \textit{open-set} setting for domain adaptation \cite{ATI}. Their solution is to iteratively associate target samples with potential labels and map samples from the source domain to target domain, in which the unknown class \rev{is} regarded as a normal class. \shh{\shhh{More} recently, the proposed OpenBP \cite{OpenBP} \shhh{adversarially} learn\shhh{s} a decision boundary for the unknown class. In contrast to these methods, we explicitly construct large margins between the known and unknown class samples and in the meantime ensure that the unknown-class samples are not aligned between domains. Therefore, we can detect unknown-class samples more robustly.}%/***Jason: need to edit the last sentence***/ }

    \noindent{\textbf{Multi-task Learning.}} Our collaborative distribution alignment is a bilateral transfer, which is related to multi-task learning \cite{zhang2017survey}. 
    Through multi-task learning, tasks collaborate by sharing \textit{parameters} \cite{AML, MRN, misra2016cross} \shh{or \textit{features} \cite{Convex, DAMFL}}, while in our method, domain collaboration is achieved by directly sharing \textit{data}. Further, one may simply tackle the weakly supervised open-set domain adaptation problem
    %construct our problem 
    with two tasks, each of which aims at training an independent classifier for a single domain. However, as illustrated in Figure \ref{fig:reid}, without the sharing of cross-domain data, many samples cannot be properly labeled. On the other hand, if we train a single classifier with shared data, the problem degenerates to single-task learning.
    \section{Collaborative Distribution Alignment}

    \subsection{Approach Overview}
    %\jie{

    To address the weakly supervised open-set domain adaptation problem, we aim to learn a \shh{shared} latent space in which samples from different domains %can be well classified. 
    \shh{are well aligned}.
    Suppose that we have two different domains $\mathcal{D_A}$ and $\mathcal{D_B}$, both of which have partially labeled samples.
    Let $\{\mathcal{D}_i = \mathcal{L}_i\bigcup\mathcal{U}_i\}_{i=A,B}$, where $\mathcal{L}_i = \{(x_i^j, y_i^j)\}_{j=1}^{l_i} $ denotes the labeled set with sample \shhh{$x_i^j\in \mathbb{R}^d$} and label $y_i^j$, and $\mathcal{U}_i = \{(x_i^j)\}_{j=l_i+1}^{l_i+u_i} $ is the unlabeled portion \shh{with unlabeled samples.} 
    Here, $x$ is the \rev{$d$-dimensional feature representation} of each \rev{image} sample, while $l_i$ and $u_i$ represent the sizes of $\mathcal{L}_i$ and $\mathcal{U}_i$, respectively. In practice, we have $l_i << u_i$.
    Given a set $\mathcal{C}_i$ in labeled portion $\mathcal{L}_i$, which includes $\shhh{|\mathcal{C}_i|}-1$ known classes and an additional unknown class that gathers instances from \shh{all the} classes that cannot be found in the known label space,
    our goal is to label each sample in $\mathcal{U}_i$ \rev{with} a \shh{known} class \shh{or detect it as an unknown-class sample}.
    
    From Figure \ref{fig:method}(a), we can see that some samples in $\mathcal{U}_A$ do not belong to $\mathcal{C}_A$ but \shhh{to} $\mathcal{C}_B$. \rev{For example, there are unlabeled square samples in Domain A, but no labeled square sample exists in this domain.} Merely using $\mathcal{L_A}$, we cannot properly label these samples.
    We then leverage the total labeled set $\mathcal{L}$, which consists of the labeled set $\mathcal{L}_A$ and $\mathcal{L}_B$ 
    with the classes $\mathcal{C_L} = \mathcal{C_A} \bigcup \mathcal{C_B}$,   
    to label all the samples in $\mathcal{U}_A$ and $\mathcal{U}_B$ 
    as one of the classes in $\mathcal{C_L}$. However, due to domain shift between $\mathcal{D_A}$ and $\mathcal{D_B}$, direct\shhh{ly} using $\mathcal{L}$ would cause significant \shh{performance degeneration}  \cite{transfer_survey}. 
% Quality control editor: Please be consistent in whether or not a space is used before a citation.    
    To solve this problem under our proposed setting, we focused on the following two most important features of our setting, which have not been well studied before: 

    1) Under the weakly supervised setting, we need to transfer knowledge bilaterally. Thus, for the first time, we introduce \textit{dual} mapping (learning dual domain-specific mappings) into the domain adaptation area. Dual mapping can better exploit domain-specific information and align both domains onto the shared latent space (see Table \ref{table:outlier}).

    2) Under the open-set setting, there could be a large number of unknown-class samples. 
    % \shhh{\sout{If we regard the unknown class as a normal category (as other domain adaptation works do), the confusion of known and unknown samples would largely undermine the overall performance.}} 
    To avoid confusing these samples with known-class ones, we are the first in domain adaptation to explicitly form large margins between known-class and unknown-class samples (Eq.\ref{d_selection}). 
    % \shhh{In addition}, unlike other methods,
    % \shhhh{\sout{domain adaptation} methods,}
    
    % we only align features of known\shhh{-class} but not unknown\shhh{-class} samples to avoid improper alignment (Eqs. 5, 6).

    %  Thus, we propose to iteratively learn \shh{two} domain-specific transform matrices $W_A$ and $W_B$ for the two domains $\mathcal{D_A}$ and $\mathcal{D_B}$, respectively, so that they are projected onto a \shh{shared} latent space. In this way, we can preserve domain-specific information as well as exploit domain-shared features. 

    In general, our method works as follows. In each iteration, we first assign pseudo-labels to a subset of samples in $\mathcal{U_A}$ and $\mathcal{U_B}$. Then, we use the estimated pseudo-labels as well as the labeled set $\mathcal{L}$ to learn optimized values $W_A'$ and $W_B'$. Next, $\mathcal{D_A}W_A'$ and $\mathcal{D_B}W_B'$ are\footnote{$\mathcal{D_A}W_A'$ means that each sample in $\mathcal{D_A}$ will be projected by $W_A'$.} used to update the pseudo-labels, and a new iteration starts. This circle continues until it converges. After the termination, any base classifiers (SVM, k-NN) can be trained on $\mathcal{L_A}W_A$ and $\mathcal{L_B}W_B$ and predict labels for $\mathcal{U_A}W_A$ and $\mathcal{U_B}W_B$. 
    \rev{We expect} that the domain shift has been largely alleviated by performing this iterative optimization. The whole process is shown in Figure \ref{fig:method} and detailed in the following sections.
    \subsection{Pseudo-Label Assignment}
    \noindent{\textbf{Label Prediction.}} In the real world, there are usually much fewer labeled samples than unlabeled ones. With the scarce labeled samples, we are not able to robustly perform the subsequent \rev{optimization of $W_A$ and $W_B$}. Therefore, \shh{to enlarge the label information set, we propose to first estimate the pseudo-labels of samples from $\mathcal{U}_A$ and $\mathcal{U}_B$}. This can be easily done by applying base classifiers trained on $\mathcal{L}$ to predict labels of $\mathcal{U}_A$ and $\mathcal{U}_B$. This base classifier can be either k-NN, standard or variants of SVM, or neural networks. In our method, we simply utilize the standard linear SVM (LSVM) with probability estimation in the \textit{libsvm} package \cite{libsvm} due to its wide use in domain adaptation works \cite{ATI,li2017prediction,CORAL,yan2018semi}.\\
    \noindent{\textbf{Outlier Detection.}}
    However, due to domain discrepancy, some of the pseudo-labels are assigned uncertainly, and therefore, false estimation would result. We need to eliminate those uncertain labels from the \rev{optimization} step in order to avoid iterative propagation of false estimations.  
% Quality control editor: It is unclear what "them" refers to in the previous sentence. Please consider replacing it with "those samples" or "those labels" or a more appropriate phrase.
        As we have gained probability distribution estimations for each sample from LSVM, \shh{we propose} to leverage \textit{information entropy} $H$ to estimate the certainty of the prediction for each sample. A higher $H$ means that the probability distribution is more sparse, indicating that this prediction is more likely to be false. Therefore, we tend to only use samples with relatively low information entropy in the mapping step. Let $x_i$ be the $i^{th}$ sample and \shhh{$C$ be the total number of classes}, and $\textbf{Y}_{ij}$ is the probability of $x_i$ belonging to class $j$, given by LSVM. The information entropy of this prediction is denoted as
        \begin{equation}
        H(x_i) = - \sum _{j=1} ^ {C}  \textbf{Y}_{ij} log( \textbf{Y}_{ij}).
        \label{out}
        \end{equation}
        Then, if $H(x_i) \geq \gamma$, we mark $x_i$ as an outlier that will not be used for dual mapping; otherwise, $x_i$ will be used. Here, $\gamma$ is an auto-adaptive threshold, which is set to the mean value of $H$ over all the samples. In the next iteration, all the unlabeled samples will be regarded equally again. Please note that unknown-class samples that are not close to any of the labeled samples would more likely be detected by this process for 
        %its
        \shh{their}
        low confidence. 

      \subsection{Dual Mapping Under Open-set Condition}
      We intend to perform the dual mapping \shh{process} by learning $W_A, W_B \in \shhh{\mathbb{R}}^{d \times d}$ for $\mathcal{D_A}$ and $\mathcal{D_B}$, respectively. It is expected that $W_A$ and $W_B$ are able to map $\mathcal{D_A}$ and $\mathcal{D_B}$ onto a \shh{shared latent space}.
      \shh{By learning these domain-specific matrices, we can better exploit domain-specific information for the two domains. This is different from most domain adaptation methods that only learn a single transform to transfer samples between domains \cite{ATI,JDA,TCA,CORAL}.} In Section \ref{dualM}, we have also empirically found that learning domain-specific transformation is more flexible for solving the weakly supervised open-set domain adaptation, \shhh{as both domains play equal roles in this problem.}

    \vspace{0.1cm}

    %\subsubsection{Known-Unknown Class Separation} 
    \noindent{\textbf{Known-Unknown Class Separation.}}
    Under the open-set setting, a main task is to detect and separate the unknown-class samples \cite{ATI}. 
    If there is no robust separation between known and unknown class samples, the overlapping of the known and unknown class samples would probably \shhh{cause the problem of} labelling a known-class sample as \shh{an} unknown impostor, or vice versa. This significantly degenerates \shhh{the overall performance}.
    To solve this problem, we propose to encourage each known-class sample to be closer to its class center than to any of the unknown-class samples by a large \textit{margin}. 

    \rev{To begin, we first compute the distances of each known-unknown sample pair in the transformed space. For each known-class sample $x$ in domain $D$ $\in$ \{$A$, $B$\}, we find its nearest unknown-class neighbor $x_u$, which is the unknown-class sample with the smallest distance from $x$. We formally define this distance as $f_u (x) = \left \Vert x W_D - x_u W_{D'} \right \Vert ^ 2 _F$, where $D'$ is the domain that $x_u$ belongs to.}
        \shh{
          Similarly, we define the distance between $x$ and its class center $\overline{x}_D^c$ as $f_c(x) = \left \Vert (x - \overline{x}_D^c ) W_D\right \Vert^2_F$. We hope that for every $x$, we have 
        \begin {equation}
        f_c(x) + 1 < f_u(x),
        \end {equation}
        where 1 is the margin parameter that sets the scale of $W_A$ and $W_B$ \cite{LMNN}. \rev{We define $\mathcal{C}_K$ as the set of all the known classes and $\textbf{X}_D^c$ as the sample set of class $c$ in domain $D$. Then, the} loss function for unknown separation in domain $D$ can be defined as 
        \begin{equation}
        \label{d_selection}
        \begin{aligned}
        U_D & = \sum \limits_{c \in \mathcal{C}_K, x^i \in \textbf{X}_D^c} \left[ 1 + f_c(x^i) - f_u(x^i) \right]_{+}, \\
        \end{aligned}
        \end{equation}}where the term $[x]_+=\max\{x, 0\}$ is the standard hinge loss, which allows us to only penalize the unknown impostors that are inside the safety margin of $x$ and $\overline{x}_D^c$. For unknown-class samples that are far enough, the hinge loss becomes $0$ and does not contribute to $U_D$. The total unknown separation loss of \rev{$\mathcal{D_A}$} and \rev{$\mathcal{D_B}$} is defined as
        \begin {equation}
        \begin{aligned}
        U = \frac{1}{2} ( U_A + U_B ).
        \end{aligned}
        \label{U}
        \end{equation}

    %\subsubsection{Aligning Known Class Samples between Domains}
    \noindent{\textbf{Aligning Known-class Samples between Domains.}}
    Under the open-set setting, aligning all samples between two domains is unnecessary because the purpose of dual mapping is to align all the known-class samples to gain a better classification performance while dragging all the unknown-class samples 
    %(i.e. outliers)
    away from the known ones to make a clear separation. 

    To this end, we discard all the unknown class samples in the distribution alignment process for domain adaptation. To measure the domain discrepancy, we adopt the \textit{maximum mean discrepancy} (MMD) \cite{MMD}, which is a widely used nonparametric domain discrepancy measurement in domain adaptation works \cite{DAN,JDA,TCA, DDC}.

    More specifically, the marginal distribution distance between the transformed domains can be computed by
          \begin{equation}
          Dist_{M} = \frac{1}{2} \left \Vert P_A W_A - P_B W_B \right \Vert ^2 _F,
          \end{equation}
    where $P_D = \frac{1}{n^k_D} \sum_{i=1}^{n_D^k}x_D^i$ is the mean of \textit{known} samples in domain $D$, and $n_D^k$ denotes the number of those samples.
          Similarly, the distance between conditional distributions $Dist_{C}$ can be measured by the sum of the distances between the corresponding \textit{known} class centers in both domains. That is defined as      
          \begin{equation}
          Dist_{C} = \frac{1}{2} \sum_{c \in \mathcal{C}_K} \left \Vert P_A^c W_A - P_B^c W_B\right \Vert ^2 _F,
          \label{dist_c}
          \end{equation}
    where $P_{D}^c = \frac{1}{n_D^c} \sum_{i=1}^{n_D^c} x_D^{c,i}$ is the mean of class $c$ \shh{samples}, and $x_D^{c,i}$ is the $i^{th}$ sample of class $c$ in domain $D$. 
    The \shh{label} of each sample is either \shh{provided by the dataset}
    %labeled 
    or assigned by pseudo-label prediction. %/***Jason: the last sentence is not properly stated? what do you mean?***/. 

    By minimizing $Dist_M$ and $Dist_C$, the discrepancies between both marginal and conditional distributions of known-class samples in \rev{$\mathcal{D_A}$} and \rev{$\mathcal{D_B}$} are reduced, while the unknown-class samples will not be improperly aligned.

    \vspace{0.1cm}

    %\subsubsection{Aggregating Same-class Samples}
    \noindent{\textbf{Aggregating Same-class Samples.}}
    To facilitate the center-based separation between known and unknown class samples \shhh{(Eq.\ref{U})} as well as the alignment between known class samples based on MMD \shhh{(Eq.\ref{dist_c})},
    we propose to explicitly \textit{aggregate} all the samples to their class centers.

    In detail, we define $\overline{x}_D^c$ as the estimated center of class $c$ in domain $D$ and $\textbf{X}_D^c$ as the sample set of class $c$ with size $n_D^c$. Then, the aggregation loss function $G_D$ can be expressed by
          \begin{equation}
          \begin{aligned} 
          G_D = \sum_{c \in \mathcal{C_L}} &\frac{1} {n_D^c} \sum_{x^i \in \textbf{X}_D^c} \left \Vert (x^i - \overline{x}_D^c)  W_\shh{D} \right \Vert^2 _F.
          \end{aligned}
          \end{equation}
          Overall, the total loss $G$ of \rev{$\mathcal{D_A}$} and \rev{$\mathcal{D_B}$} is then defined as
          \begin{equation}
          \begin{aligned} 
          G = \frac{1}{2} (G_A + G_B).
          \end{aligned}
          \end{equation}

    \noindent{\textbf{Objective Function.}}
    With all the components discussed above, our method can be formed now. To configure the strength of each component, we introduce balancing parameters $\lambda_M, \lambda_G,\text{ and }\lambda_U$ for marginal distribution alignment, center aggregation and known-unknown class separation, respectively. We define the total loss function as 
          \begin{equation}
          \begin{aligned}
          \label{total_equation}
          f = Dist_C + \lambda_M Dist_M + \lambda_G G + \lambda_U U
          \end{aligned}.
          \end{equation}      
          Then, our objective function can be concluded as follows:
          \begin{equation}
          \label{opt_task}
          (W_A^{*}, W_B^{*}) = \mathop{\arg\min}_{W_A, W_B} f (\mathcal{D_A}, \mathcal{D_B}, W_A, W_B).
          \end{equation}

          Details of the optimization process are given in the supplementary material\footnote{\url{https://ariostgx.github.io/website/data/CVPR_2019_CDA_supp.pdf}} due to lack of space.

          % The loss function $f$ is not convex in the elements of $W_A$ and $W_B$. Therefore, we use the non-linear methods in  \cite{svanberg2002class} to optimize the transform matrices. Due to the lack of space, we include further computational details in the supplementary material.
          
          % We jointly optimize $W_A$ and $W_B$ by concatenating them as a whole matrix $W \in \shhh{\mathbb{R}} ^{d \times 2d}$. By optimizing $W$ as a whole, we can ensure that both matrices are simultaneously updated.
          % %\shh{\sout{, and thus can project $\mathcal{D_A}$ and $\mathcal{D_B}$ \shh{onto} a \shh{shared} latent domain.}}
          % As for gradient $\partial{f} / \partial{W}$, we similarly concatenate $\partial{f} / \partial{W_A}$ and $\partial{f} / \partial{W_B}$ to form a single gradient value. After the optimization of $W_A$ and $W_B$, we can easily obtain the projected features of \rev{$\mathcal{D_A}$} and \rev{$\mathcal{D_B}$} from $\mathcal{D_A} W_A$ and $\mathcal{D_B} W_B$. Due to the lack of space, all the gradients are computed in the supplementary material.

    \vspace{-0.2cm}

    \section{Experiments}  

    \subsection{Office Dataset} \label{exp_office}

    \textit{The Office} dataset \cite{OFFICE} is a standard benchmark for domain adaptation evaluation \cite{ATI,DAN,CORAL,OpenBP} and consists of three real-word object domains: \textit{Amazon} (A), \textit{Webcam} (W) and \textit{DSLR} (D). In total, it has 4,652 images with 31 classes.
    We construct three dual collaboration tasks: A $\leftrightarrow$ W, A $\leftrightarrow$ D and W $\leftrightarrow$ D, where $\leftrightarrow$ denotes the collaboration relation between two domains. Similar to  \cite{ATI}, we extract the features using ResNet-50 \cite{ResNet} \shh{pretrained} on ImageNet \cite{ImageNet}. For parameters, we set $\lambda_M = 10, \lambda_G = 1$ and $\lambda_U = 0.1$.

    \noindent {\bf Protocol.} We randomly selected 15 classes as the known classes, and the remaining classes are set as one unknown class, \textit{i.e.}, there are 16 classes in total. 
    Both domains contain all 16 classes, while only a subset of them have labeled samples.
    Specifically, we randomly selected 10 known classes to form the  \shhh{labeled known-class \rev{set}} for each domain, \shhh{with the constraint that} 5 classes are shared between them. Similar to  \cite{ATI}, for each domain, we randomly took 3 samples from each \rev{class in the} \shhh{labeled known-class \rev{set}} and 9 samples from the unknown class, as labeled samples, and set all the remaining as unlabeled. We expect each unlabeled sample to be correctly labeled either as one of the 15 known classes or the unknown class. The above is repeated 5 times, and the mean accuracy and standard deviation of each method are reported.
% Editor: Please ensure that the intended meaning has been maintained in the edit of the previous sentence.

    \noindent {\bf Compared Methods.} We first compared the proposed CDA with two open-set domain adaptation methods, ATI \cite{ATI} and OpenBP \cite{OpenBP}. The conventional unsupervised domain \shhh{adaptation} methods, including TCA \cite{TCA}, GFK \cite{GFK}, CORAL \cite{CORAL}, JAN \cite{JAN} and PADA \cite{PADA}, are compared as well. \shhh{Since our method learns dual projections, to undertake a fair comparison, for methods that require a single labeled domain (ATI, OpenBP, JAN, and PADA), we apply them by two steps: given $\mathcal{L_A}$ as the source and $\mathcal{D_B}$ as the unlabeled target, we learn to transform $\mathcal{L_A}$ to $\mathcal{D_B}$, and label $\mathcal{U_B}$ with $\mathcal{L_A}$ and $\mathcal{L_B}$. Then, we learn to label $\mathcal{U_A}$ in a symmetrical way. For other methods, we simply transform $\mathcal{D_A}$ to $\mathcal{D_B}$.}

    Since there are a small number of labeled samples in the target domain for semi-supervised domain adaptation, we also compared two semi-supervised methods, namely MMDT \cite{MMDT} and ATI-semi \cite{ATI}. \shhh{We test these methods in a way similar to methods such as ATI and JAN, except the target domain in each symmetry step here is semi-supervised.} 

    In addition, we compared multi-task learning methods since our method benefits from the collaborative transfer between two domains, which \shhh{is} related to multi-task learning. For compariso\shhh{n}, we report the results of CLMT \cite{Convex}, AMTL \cite{AML}, and MRN \cite{MRN}.
    By following  \cite{Convex, AML}, every task in CLMT and AMTL is set as a one-vs-all binary classification problem. For MRN, we denote two tasks as learning classifiers on $\mathcal{D_A}$ and $\mathcal{D_B}$, separately. 

    The \shh{results} of the above methods are obtained by linear SVM and trained and tested on transformed features, except for CLMT and AMTL, which directly learn \shh{classifiers}. As baselines, we report results of the LSVM trained \shh{with non-adaptation data $\mathcal{L}$}, which we call \shh{NA}. We also report another baseline called \shh{NA-avg}, which is the average accuracy of training and testing a LSVM for each domain independently. \shhh{Moreover, we have also undertaken a comparison with methods in  \cite{SA, RevGrad, DAN, JDA, DCORAL, JGSA, DDC}, the results of which are less competitive and are thus only included in the supplementary material for sake of space.}
   
 \begin{table} [!ht]%\footnotesize
                \renewcommand{\arraystretch}{0.8}
                \vspace{-0.2cm}
                \centering
                  {\scriptsize
                  \begin{tabular}
                        {p{1.3cm}<{\centering}|*{3}{p{1.2cm}<{\centering}|}{p{0.9cm}<{\centering}}}
                          \hline
                          Methods & A $\leftrightarrow$ W & A $\leftrightarrow$ D & W $\leftrightarrow$ D & AVG. \\
                          \hline
                          NA-avg & 62.2 $\pm$  3.19 & 61.0 $\pm$  3.05 & 66.3 $\pm$ 1.72 & 63.12 \\
              NA & 72.6 $\pm$ 2.04 & 69.4 $\pm$ 2.22 &  84.2 $\pm$ 3.89 & 75.40 \\ 
                          \hline
                          TCA \cite{TCA} & 73.3 $\pm$ 1.67 & 70.6 $\pm$ 1.99 & 83.5 $\pm$ 3.96 & \textbf{\color{blue}75.80} \\
                          GFK \cite{GFK} & 56.9 $\pm$ 2.89 & 55.7 $\pm$ 1.46 & 70.9 $\pm$ 4.85 & 61.17 \\
                          CORAL \cite{CORAL} & 69.9 $\pm$ 4.41 & 67.7 $\pm$ 3.13 & 83.7 $\pm$ 3.65 & 
                          73.77 \\
                          JAN \cite{JAN} & 63.8 $\pm$ 1.27 & 65.5 $\pm$ 0.76 & 74.7 $\pm$ 1.41 & 68.00 \\
                          PADA \cite{PADA} & 60.3 $\pm$ 1.09 & 60.2 $\pm$ 0.98 & 70.9 $\pm$ 1.88 & 63.80 \\
                          ATI \cite{ATI} & 70.4 $\pm$ 4.15 & 65.9 $\pm$ 1.80 & 81.7 $\pm$ 4.74 & 72.67 \\
                          OpenBP \cite{OpenBP} & 62.6 $\pm$ 4.11 & 62.9 $\pm$ 1.71 &  67.9 $\pm$ 2.31 &  64.47 \\
                          \hline
                          ATI-semi \cite{ATI} & \textbf{\color{blue}73.4 $\pm$ 2.28} & \textbf{\color{blue}72.0 $\pm$ 2.87} & 77.8 $\pm$ 3.40 & 74.72 \\   
                          MMDT \cite{MMDT} & 72.6 $\pm$ 2.15 & 69.4 $\pm$ 2.19 & \textbf{\color{blue}84.4 $\pm$ 3.99} & 75.47 \\
                          \hline 

                          AMTL \cite{AML} & 50.2 $\pm$ 1.45 & 48.8 $\pm$ 0.90 & 62.1 $\pm$ 2.17 & 53.70 \\
                          CLMT \cite{Convex} & 50.3 $\pm$ 1.47 & 50.0 $\pm$ 0.77 & 61.7 $\pm$ 1.23 & 54.00 \\
                          MRN \cite{MRN} & 62.2 $\pm$ 3.38 & 62.4 $\pm$ 2.44 & 77.4 $\pm$ 3.43 & 67.33\\
                          \hline
                          \bf CDA & \textbf{\color{red}77.1 $\pm$ 1.35} & \textbf{\color{red}75.2 $\pm$ 1.63} & \textbf{\color{red}88.1 $\pm$ 2.45} & \textbf{\color{red}80.13} \\
                          \hline 
                    \end{tabular}}
                    \caption{%\footnotesize
                      %\small
                  Comparing state-of-the-art methods on \textit{Office}. The $1^\text{st}/2^\text{nd}$ best results are indicated in {\color{red}red}/{\color{blue}blue}.
                  }
                  \label{table:office_3}
                    \vspace{-0.2cm}
                    
                  \end{table}

 \begin{table*} [!htb]
    %\begin{table*} [h]
    \renewcommand{\arraystretch}{0.8}
      \centering
      {\scriptsize
        \begin{tabular}
          {p{1.40cm}<{\centering}|*{17}{p{0.4cm}<{\centering}|}{p{0.4cm}<{\centering}}}
          \hline
          \multirow{2}{*}{Method} 
          &  \multicolumn{2}{c|}{1 $\leftrightarrow$ 2} &  \multicolumn{2}{c|}{2 $\leftrightarrow$ 3} &  \multicolumn{2}{c|}{3 $\leftrightarrow$ 4} &  \multicolumn{2}{c|}{4 $\leftrightarrow$ 5} &  \multicolumn{2}{c|}{5 $\leftrightarrow$ 6} &  \multicolumn{2}{c|}{6 $\leftrightarrow$ 7} &  \multicolumn{2}{c|}{7 $\leftrightarrow$ 8} &  \multicolumn{2}{c|}{8 $\leftrightarrow$ 1} &  \multicolumn{2}{c}{AVG.} \\
          \cline{2-19}
          & r=1 & r=5 & r=1 & r=5 & r=1 & r=5 & r=1 & r=5 & r=1 & r=5 & r=1 & r=5 & r=1 & r=5 & r=1 & r=5 & r=1 & r=5 \\
          \hline
          %\hline
          NA-avg & 43.6  & 55.3  & 29.6  & 38.2  & 50.1  & 54.8  & 69.2  & 95.2  & 38.2  & 54.4  & 33.5  & 42.4  & 22.0  & 40.3  & 76.7  & 95.4  & 45.4  & 59.5 \\
          %\hline
          NA & 59.3  & 75.1  & 41.4  & 53.2  & 64.4  & 78.8  & 74.1  & 98.9  & 49.0  & 67.1  & 51.9  & 65.4  & 27.6  & 49.6  & 78.3  & 98.9 & 55.7  & 73.4 \\
      \hline
          %\hline
          TCA \cite{TCA} & 58.6  & 74.7  & 41.0  & 53.7  & 64.3  &  \textbf{\color{blue}79.0}  & 75.3  & 98.9 & 48.6  & 67.3  & 51.5  & 65.3  & 27.2  & 49.8  & 78.3  & \textbf{\color{blue}98.9}  & 55.6  & 73.5  \\
          GFK \cite{GFK} & 59.1  & 75.1  & 41.5  & 53.5  & \textbf{\color{blue}64.5}  &  79.0  & 75.3  & \textcolor[rgb]{ 1,  0,  0}{\textbf{99.0}} & 49.0  & 67.4  & 51.9  & 65.5  & 27.7  & 50.3  & 78.5  & 98.9  & 55.9  & 73.6  \\

          CORAL \cite{CORAL} & 58.9  & 75.8  & 41.6  & 54.2  & 64.2  & 78.4  & 74.5  & 98.8  & 47.7  & 66.3  & 51.5  & 64.6  & 26.5  & 48.6  & 78.4  & 98.8  & 55.4  & 73.2  \\
          %\hline

          JAN \cite{JAN} & 24.0  & 43.7  & 34.4  & 64.4  & 21.4  & 38.6  & 75.5  & 90.2  & 30.2  & 60.0  & 27.7  & 52.4  & 44.6  & 72.1  & 81.7  & 92.5  & 42.4  & 64.2  \\
          PADA \cite{PADA} & 14.0  & 30.7  & 39.4  & 62.2  & 22.1  & 35.3  & 75.4  & 89.0  & 30.1  & 57.4  & 27.2  & 50.4  & 47.4  & 70.9  & 77.6  & 91.1  & 41.6  & 60.9  \\
          ATI \cite{ATI} & 58.6  & 74.0  & 40.6  & 49.5  & 65.4  & 79.2  & 36.1  & 78.0  & 47.9  & 63.6  & 52.4  & 65.6  & 24.6  & 42.7  & 28.2  & 84.4  & 44.2  & 67.1  \\
          OpenBP \cite{OpenBP} & 34.7  & 49.7  & 45.9  & 66.5  & 27.3  & 39.8  & 77.1  & 90.6  & 8.1  & 26.5  & 37.6  & 53.5  & 47.5  & 67.3  & 83.5  & 93.3  & 45.2  & 54.9  \\
          \hline 
          MRN \cite{MRN} & 26.2  & 46.5  & 19.2  & 32.8  & 34.0  & 49.6  & 74.9  & 94.4  & 25.9  & 49.9  & 24.4  & 43.1  & 19.9  & 50.2  & 53.6  & 85.1  & 34.8  & 56.5  \\
          \hline %\hline
          ATI-semi \cite{ATI} & 53.4  & 71.8  & \textbf{\color{blue}55.6}  & \textbf{\color{blue}71.8}  & 56.5  & 76.8  & 39.7  & 88.9  & 55.3  & 69.3  & \textbf{\color{blue}57.6}  & \textbf{\color{blue}76.9}  & \textbf{\color{blue}46.7}  & 65.6  & 30.6  & 86.3  & 49.4  & 75.9  \\
          %\hline
          MMDT \cite{MMDT} & 36.0  & 51.1  & 35.7  & 51.5  & 57.5  & 73.2  & 74.2  & 98.9  & 29.2  & 53.2  & 22.0  & 39.5  & 19.1  & 43.5  & 76.5  & 98.6  & 42.9  & 62.3  \\
        \hline %\hline
          
          LMNN \cite{LMNN} & \textcolor[rgb]{ 0,  0,  1}{\textbf{59.4}} & \textcolor[rgb]{ 0,  0,  1}{\textbf{78.1}} & 42.5  & 59.1  & 61.8  & 74.1  & 84.7  & 96.3  & 53.3  & 72.2  & 50.8  & 66.1  & 33.5  & 60.8  & \textbf{\color{red}90.2}  & 97.7  & \textcolor[rgb]{ 0,  0,  1}{\textbf{59.5}} & 75.5  \\
          %\hline
          KISSME \cite{KISSME} & 55.2  & 68.6 & 49.2 & 71.0  & 63.1  & 75.3  & \textcolor[rgb]{ 1,  0,  0}{\textbf{86.0}} & 94.9  & \textcolor[rgb]{ 0,  0,  1}{\textbf{57.6}} & 74.5  & 55.3 & 71.6 & 45.8 & \textcolor[rgb]{ 0,  0,  1}{\textbf{71.6}} & 22.0  & 56.7  & 54.3  & 73.0  \\
          %\hline
          XQDA \cite{XQDA} & 55.6  & 68.9  & 49.1  & 71.0 & 63.1  & 75.3  & \textcolor[rgb]{ 0,  0,  1}{\textbf{86.0}} & 94.9  & 57.6  & \textcolor[rgb]{ 0,  0,  1}{\textbf{74.5}} & 55.3  & 71.6  & 45.8  & 71.6  & 48.2  & 86.0  & 57.6  & \textcolor[rgb]{ 0,  0,  1}{\textbf{76.7}} \\
          %\hline
          DLLR \cite{DLLR} & 59.2  & 75.0  & 41.3  & 52.9  & 64.1  & 78.1  & 72.9  & 98.7  & 48.7  & 66.9  & 51.8  & 64.5  & 27.9  & 49.6  & 78.5  & \textcolor[rgb]{ 1,  0,  0}{\textbf{99.0}} & 55.6  & 73.1  \\
          SPGAN \cite{Deng_2018_CVPR} & 41.6  & 59.9  & 31.3  & 47.5  & 44.7  & 58.6  & 69.6  & 97.7  & 41.4  & 61.0  & 43.3  & 59.6  & 22.8  & 46.3  & 74.1  & 98.0  & 48.7  & 69.6  \\
          \hline %\hline
          \textbf{CDA} & \textcolor[rgb]{ 1,  0,  0}{\textbf{64.6}} & \textcolor[rgb]{ 1,  0,  0}{\textbf{80.4}} & \textcolor[rgb]{ 1,  0,  0}{\textbf{62.4}} & \textcolor[rgb]{ 1,  0,  0}{\textbf{88.1}} & \textcolor[rgb]{ 1,  0,  0}{\textbf{67.9}} & \textcolor[rgb]{ 1,  0,  0}{\textbf{84.9}} & 76.0  & \textcolor[rgb]{ 0,  0,  1}{\textbf{98.9}} & \textcolor[rgb]{ 1,  0,  0}{\textbf{61.5}} & \textcolor[rgb]{ 1,  0,  0}{\textbf{82.1}} & \textcolor[rgb]{ 1,  0,  0}{\textbf{59.6}} & \textcolor[rgb]{ 1,  0,  0}{\textbf{78.1}} & \textcolor[rgb]{ 1,  0,  0}{\textbf{67.2}} & \textcolor[rgb]{ 1,  0,  0}{\textbf{83.8}} & \textbf{\color{blue}85.6}  & 98.5  & \textcolor[rgb]{ 1,  0,  0}{\textbf{68.1}} & \textcolor[rgb]{ 1,  0,  0}{\textbf{86.9}} \\

          \hline
      \end{tabular}}
      \caption{
        %\footnotesize
            %\small
        Comparing CMC accuracies with state-of-the-art methods on DukeMTMC-reID (\%).
        } 
        \label{table:REID}
      \vspace{-0.3cm}
    \end{table*}

    \noindent \textbf{Comparison Results.} The results are shown in Table \ref{table:office_3}. We can observe that our proposed CDA method outperforms all the \shh{other methods} in the three tasks. For example, CDA respectively surpasses the best alternative TCA by 3.8\%, 4.6\% and 4.6\% in A$\leftrightarrow$W, A$\leftrightarrow$D and W$\leftrightarrow$D, respectively. Further, CDA exceeds \shh{TCA} by 4.33\% (80.13\%-75.80\%) \shh{on average}.
    %\shh{\sout{in mean accuracy}}
    The reasons may be as follows. (1) Unlike \shh{closed-set} methods (such as TCA) that align distributions over all the samples, our CDA is well designed to only align feature distributions of the known classes. By doing so, we avoid a negative transfer caused by aligning distributions of the unknown class, the samples of which may actually come from different classes with inherently different features. (2) CDA learns more discriminative
    features by minimizing intraclass variation and explicitly separating known and unknown class samples, \shh{while most other methods only focus on learning domain-invariant features. (3) CDA simultaneously exploits unlabeled information in both domains, while other methods \shhh{that require a fully labeled source domain} (such as ATI) only use unlabeled data in a single domain in each symmetry step.} 

    Additionally, our method outperformed semi-supervised domain adaptation methods. This could be because we simultaneously utilize unlabeled data from both domains, while these methods use them separately. Moreover, CDA largely exceeds multi-task learning methods, which occurs mainly because (1) CDA \shh{aligns both domains}, but \shh{AMTL and CLMT ignore the domain discrepancy in training samples and because (2) CDA merges incomplete label spaces of $\mathcal{L_A}$ and $\mathcal{L_B}$ to form a complete one, while in MRN, these label spaces \shhh{are separated}.}

     \vspace{-0.2cm} 

      \subsection{Person Re-identification (re-id) Application} \label{exp_reid}  
      \vspace{-0.1cm} 
      A natural practical example of our proposed setting can be the person re-identification (re-id) problem \cite{gong2014person}, which is widely and deeply researched in computer vision for visual surveillance \cite{Deng_2018_CVPR,DUKE2,8576568,OpenWorld}. \shhh{Re-id} aims to match people's images across nonoverlapping camera views. Currently, training state-of-the-art re-id models requires a large number of cross-camera labeled images \cite{KISSME,XQDA,LMNN},
      but it is very expensive to manually label (pairwise) people images across cameras views. \rev{Further, as mentioned before, we cannot find an off-the-shelf source domain for these people.} Thus, it is meaningful to annotate images in each camera pair by performing domain adaptation  collaboratively between these camera views. Moreover, it is hard to guarantee that people appearing in one camera view would also appear in another view. This problem introduces the \textit{open-set} setting.
      \rev{Therefore, to show the practicality of the proposed setting and the effectiveness of CDA, we conduct an experiment on a real-world re-id application in this subsection.}

      DukeMTMC-reID \cite{DUKE} \shh{is a popular re-id benchmark \cite{Deng_2018_CVPR, DUKE2, DUKE1, DUKE3}}. It consists of person images from 8 \shh{nonoverlapping} camera views. The DukeMTMC-reID dataset is a genuine \shhh{open-set} re-id dataset, since \shh{for each camera pair, there are a number of identities that only appear in one view and thus lack an image to match in another view. For example, in camera pair 1 $\leftrightarrow$ 2, there are 105 identities unique to camera 1 and 84 identities unique to camera 2. }
      For evaluation, \shhh{we conducted our experiments on the largest portion \textit{gallery}}. We constructed 8 camera pairs for our experiment, \rev{where each camera view appears twice}, and extracted features with ResNet-50 that was pretrained on Market-1501 \cite{MARKET} using the method in  \cite{IDE}. For parameters, we set $\lambda_M = 10, \lambda_G = 1$, and $\lambda_U = 1$.

    \noindent{\bf Protocol.}
    For \shhh{both} camera views, we let the classes that only appear in a single camera view belong to the unknown class, while the shared classes are set as known classes. \rev{To simulate the weakly supervised setting, in each camera view, all the known classes exist, but only a subset of them have labeled samples. We call the set of known classes that have labeled samples the \textit{labeled known-class set}.} Suppose \shhh{there} are $N$ known classes. \shhh{Then, for each camera view, we randomly select 2$N$/3 known classes \rev{to form} its labeled known-class \rev{set}, with the constraint that $N$/3 classes are shared between \rev{these sets of both views}. Then, \rev{for each view}, we randomly take 1 image as a labeled sample \rev{from each class in its labeled known-class set}. Additionally, $N$/4 samples from the unknown class are labeled as well. All the rest are set as unlabeled.}     
    During testing, for each image in the unlabeled set, we retrieve the matched persons from the labeled set. Then, we use the cumulative match characteristic (CMC) to measure the re-id matching performance. 

    \noindent {\bf Compared Methods.} We first evaluated all the methods mentioned in Section \shh{\ref{exp_office}},   
    except AMTL and CLMT, since they merely learn a classifier and are thus unsuitable for distance-based person re-id tasks. Moreover, we include common re-id oriented methods (1)LMNN \cite{LMNN}, (2)KISSME \cite{KISSME}, (3)XQDA \cite{XQDA}, and (4) DLLR \cite{DLLR}, \shhh{as well as the newly proposed SPGAN \cite{Deng_2018_CVPR}}. \shhh{To test SPGAN, for each pair, we use this model to transfer the image style of each sample in the first view to the other view and then evaluate on these transferred images.} As baselines, we report results of directly computing Euclidean distances on the original space, namely, NA and NA-avg (similar \shhh{to the} baselines used in Section \ref{exp_office}).

    \begin{figure*}[htbp]
      \centering                                      

      \subfigure[$\lambda_M$]{\begin{minipage}{4.125cm}
      \centering        
      \includegraphics[width=1\textwidth]{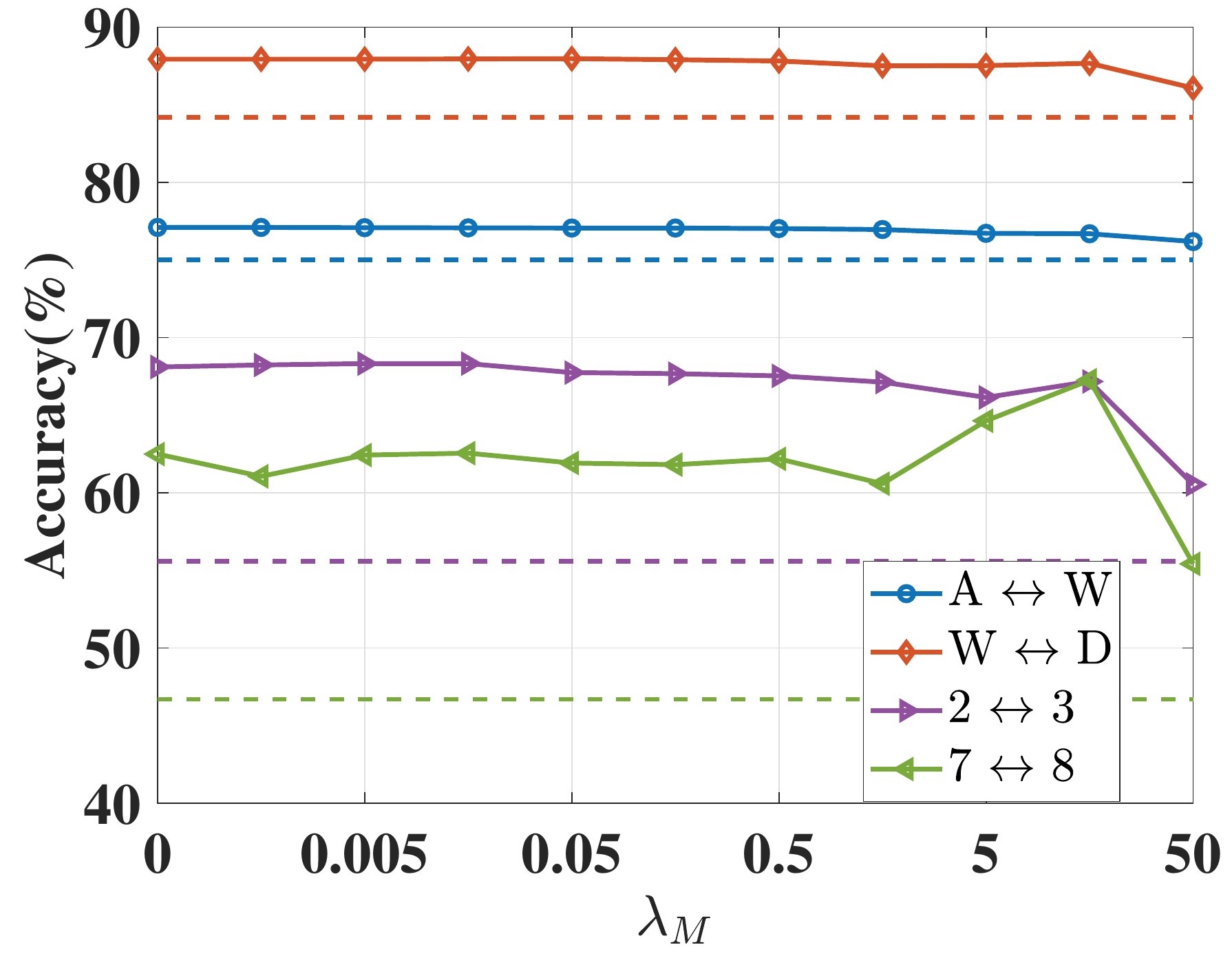} 
      % {figure/Mplot.pdf}             
      \end{minipage}}
      \subfigure[$\lambda_U$]{\begin{minipage}{4.125cm}
      \centering                                         
      \includegraphics[width=1\textwidth]{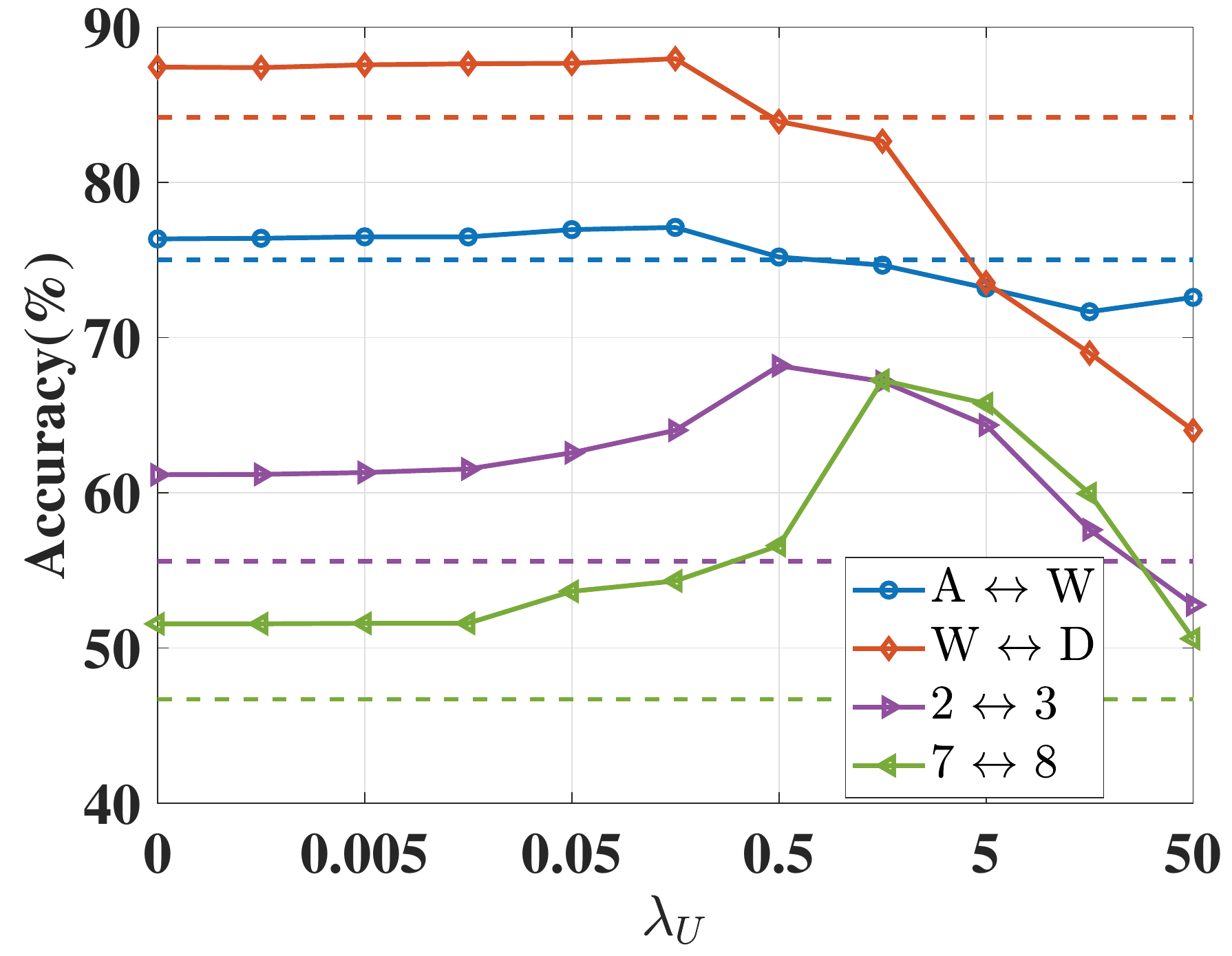} 
      %{figure/Uplot.pdf}        
      \end{minipage}}
      \subfigure[$\lambda_G$]{\begin{minipage}{4.125cm}
      \centering                                         
      \includegraphics[width=1\textwidth]{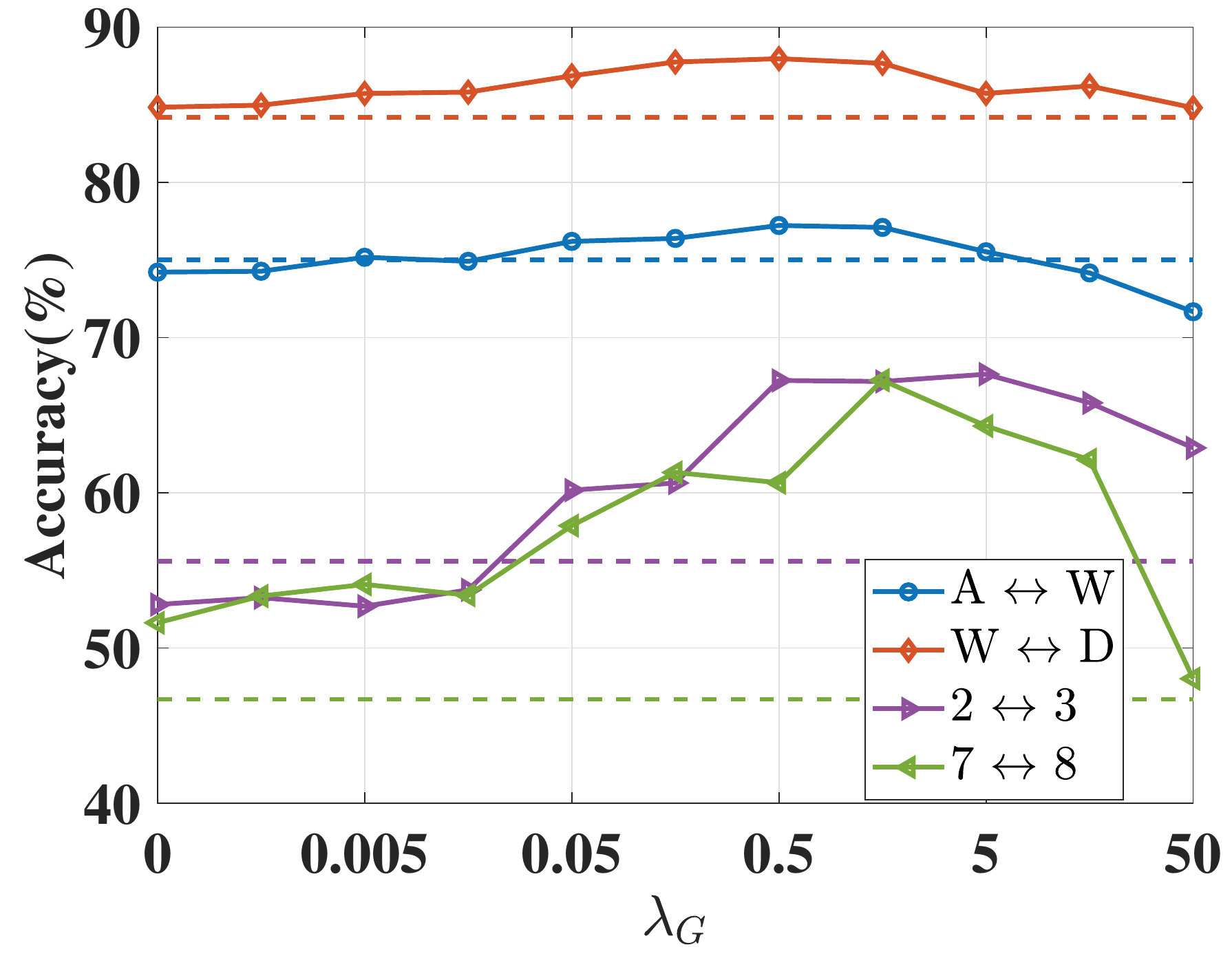}         
      \end{minipage}}   
      \subfigure[convergence]{\begin{minipage}{4.125cm} 
      \centering                                         
      \includegraphics[width=1\textwidth]{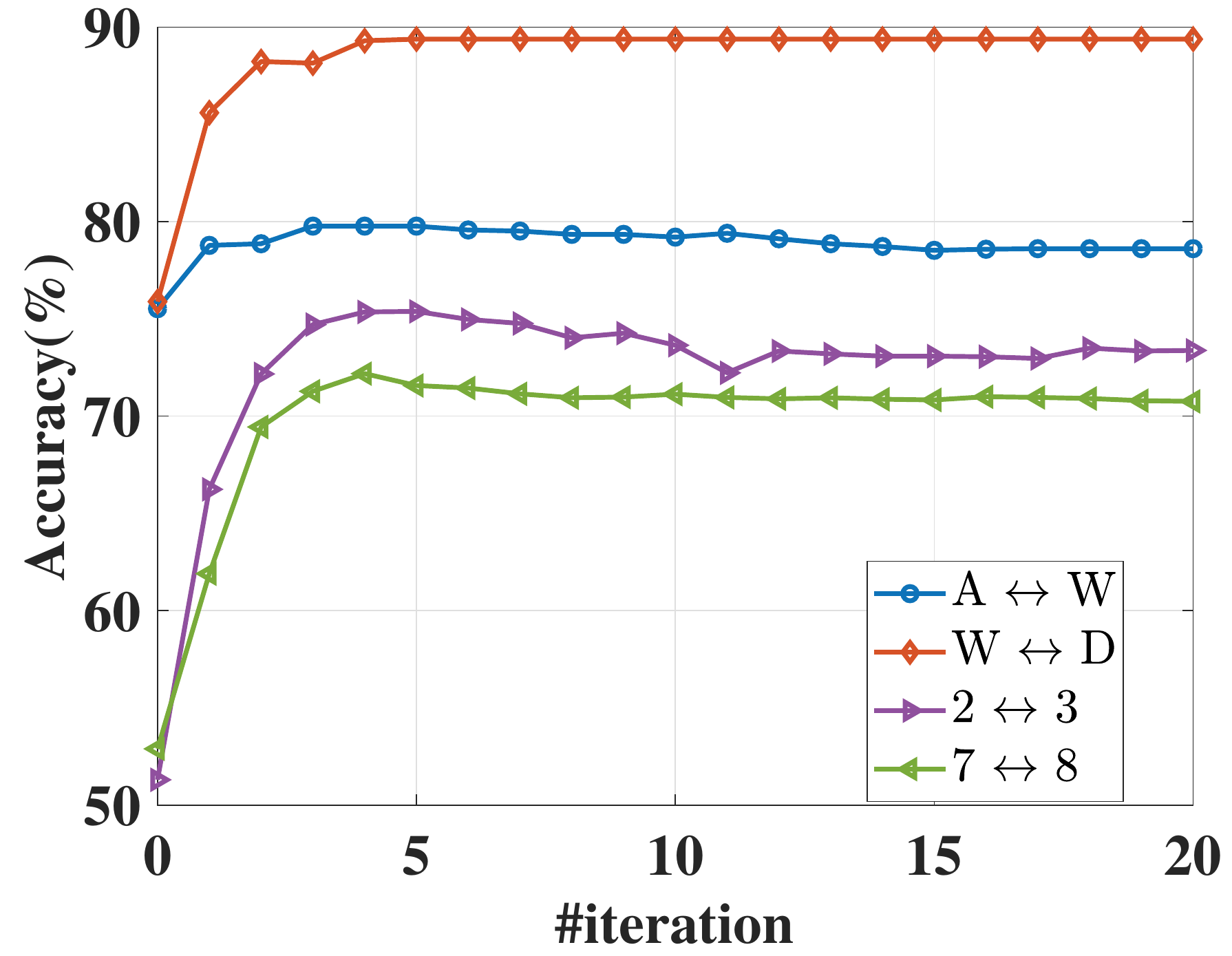}            
      \end{minipage}} 
      \caption{Classification accuracy \textit{w.r.t.} $\lambda_M$, $\lambda_U$, $\lambda_\shhh{G}$ in \textbf{(a)}, \textbf{(b)}, and \textbf{(c)}, respectively, where the dashed lines represent the best state-of-the-art performances for each task. \textbf{(d)} illustrates the convergence of the proposed method. } 
        \label{fig:sensitivity}     
        \vspace{-0.5cm}
      \end{figure*}

    \noindent {\bf Comparison Results.}
        The results are shown in Table \ref{table:REID}. Our method outperformed all the other methods in 6 of the 8 total tasks, for example, surpassing the best alternative LMNN by 5.2\%, 19.9\%, 8.2\%, and 33.7\% at rank-1 on 1$\leftrightarrow$2, 2$\leftrightarrow$3, 5$\leftrightarrow$6 and 7$\leftrightarrow$8, respectively. The performance margins over the other methods are larger still.
        On average, our method outperforms the most competitive methods by 8.6\% (68.1\%-59.5\%) and 10.2\% (86.9\%-76.7\%) at rank-1 and rank-5, respectively. 
        It is obvious that CDA significantly outperformed NA and NA-avg, while almost all the compared methods at least once performed worse than the baselines due to negative transfer.

% Editor: Please ensure that the intended meaning has been maintained in the edit of the previous sentence.
    Compared to typical methods for re-id, CDA achieves better performance mainly because (1) CDA makes use of a large number of unlabeled samples by pseudo-label assignment, while LMNN, KISS and XQDA merely rely on labeled samples, and (2) CDA explicitly separates known and unknown-class samples, while the compared methods \shhh{ (such as SPGAN)} do not distinguish these two kinds of samples.
    The superior performance of CDA in re-id further validates the effectiveness of the proposed method.

    % \vspace{-0.2cm}
    \subsection{Further Analysis of CDA}
    % \vspace{-0.2cm}
    %\subsubsection{Effect of Each component.}
    \noindent{\textbf{Effect of Each Component.}}
    % Quality control editor: Please use a consistent   capitalization format for section headings throughout the manuscript. Some journals request a specific style, so please review the journal's guidelines.
    Our loss function in \shhh{Equation} \ref{total_equation} involves four components: the marginal and conditional distribution alignments ($Dist_M$, $Dist_C$), the intraclass aggregation ($G$), and the unknown-class sample separation ($U$). We empirically evaluated the importance of each component. For each component, we report the performance of the CDA model when each component is eliminated. The results are shown in Table \ref{table:Ablation}. We can observe that the accuracy of CDA would drop once we remove any of the components. For example, on task \shh{5$\leftrightarrow$6, the performance decreased by 11.7\%, 4.2\%, 38.4\% and 1.2\%} when $G$, $U$, $Dist_C$, and $Dist_M$ was removed, respectively. 
        \begin{table} [ht]%\footnotesize 
          \renewcommand{\arraystretch}{0.8}
          \vspace{-0.2cm} 
          \centering
              {\scriptsize
              \begin{tabular}
                    {p{1.8cm}<{\centering}|*{4}{p{0.8cm}<{\centering}|}{p{0.8cm}<{\centering}}}
                      \hline
                      Models & A$\leftrightarrow$D & A$\leftrightarrow$W & W$\leftrightarrow$D & 5$\leftrightarrow$6 & 7$\leftrightarrow$8 \\
                      \hline %\hline
                      %\hline
                      Missing $G$ & 70.0  & 73.5  & 84.7 & 49.8 & 51.6\\ 
                      %\hline
                      Missing $U$ & 74.4  & 76.4  & 86.8 & 57.3 & 51.6\\
                      %\hline
                      Missing $Dist_C$ & 64.9 & 70.3 & 79.4 & 23.1 & 11.2\\
                      Missing $Dist_M$ & 74.8  & 77.1  &  87.9  & 60.3 & 62.5\\
                      \hline 
                      CDA & \textbf{75.2}  & \textbf{77.1}  & \textbf{88.1} & \textbf{61.5} & \textbf{67.2}\\
                      \hline
                \end{tabular}}
                \caption{%\footnotesize
                %\small
                Effect of components of Eq.\ref{total_equation} (\%) 
                }
                \label{table:Ablation}
                \vspace{-0.2cm}
              \end{table}

    %\subsubsection{Parameter Sensitivity}
    \noindent{\textbf{Parameter Sensitivity.}}
    %Similar to other state-of-the-art domain adaptation methods \cite{JDA,JGSA}, our method also involves several parameters. 
    In this section, we evaluated the parameter sensitivity of our model. We run CDA with a large range of parameters $\lambda_M$, $\lambda_U$ and $\lambda_\shhh{G}$ from 0 to \shh{50}. The results of randomly chosen tasks are shown in Figure \ref{fig:sensitivity} on the two datasets used in our experiments. As shown, there should be a proper range for each parameter. Taking $\lambda_U$ for example, when $\lambda_U$ is too large, we mainly learn to separate known/unknown-class samples, making feature alignment weaker; when it is too small, we are not able to adequately separate known/unknown-class samples. 
    % , making \rev{unknown-class} detection harder and overall accuracy worse. 
    Nevertheless, it can be observed that the performance of CDA is robust for $\lambda_M \in [0.001, 10]$, $\lambda_U \in [\shhh{0.01}, 1]$, and $\lambda_G \in [0.5, 10]$. 
    % In practice, one can pick values within these robust ranges to obtain decent enough performances. 
    % Specially, we suggest setting a larger $\lambda_U$ for applications where there are a larger amount of unknown classes (like re-id), aims to better separate known\shhh{-class} from unknown\shhh{-class} samples. 
    
    \noindent{\textbf{Convergence.}}
    We empirically tested the convergence property of CDA. The result in Figure \ref{fig:sensitivity} (d) shows that the accuracy increases and converges in approximately 5 iterations.  

    \noindent{\textbf{Effect of Outlier Detection.}}
    We detect and remove low-confidence pseudo-labels \shh{by} setting a threshold of information entropy in \shhh{Eq.\ref{out}}. To show the effectiveness of this process, we compared CDA with the CDA without outlier detection. As shown in the second and fourth rows of Table \ref{table:outlier}, all the tasks significantly benefit from this process, which shows \shh{the} importance of the outlier detection. 
    %\shh{\sout{, which not only produces better results, but also effectively avoids our method from turning into a vicious circle.}}
          \begin{table} [ht]%\footnotesize
             \vspace{-0.2cm}
          \renewcommand{\arraystretch}{0.8}
          \centering
              {\scriptsize
              \begin{tabular}
                    {p{1.6cm}<{\centering}|*{4}{p{0.8cm}<{\centering}|}{p{0.8cm}<{\centering}}}
                      \hline
                      Models & A$\leftrightarrow$D & A$\leftrightarrow$W & W$\leftrightarrow$D & 5 $\leftrightarrow$ 6 & 7 $\leftrightarrow$ 8 \\
                      \hline %\hline
                      no outlier & 74.3 & 73.4 & 86.0 & 52.7 & 38.0 \\
                      %\hline
                      single mapping & 70.2 & 69.8 & 81.1 & 56.5 & 33.9 \\
                      \hline
                      CDA  & \textbf{75.2} & \textbf{77.1}& \textbf{88.1} & \textbf{61.5} & \textbf{67.2} \\
                      \hline
                \end{tabular}}
                \caption{
            Effect of outlier detection and dual mapping (\%)
            }
            \label{table:outlier}
                \vspace{-0.3cm}
              \end{table}

    \label{dualM} 
    \noindent{\textbf{Effect of Dual Mapping.}}
    To map each domain to the \shh{shared} latent space, we learn a different transformation matrix for each domain. To compare this fashion with learning a shared transform matrix for both domains, we conducted CDA by learning the same transformation matrix for both domains. The results shown in Table \ref{table:outlier} demonstrate the advantage of dual mapping. This outcome is probably because dual mapping enables learning \shh{more} domain-specific \rev{information} for each domain during the adaptation \shh{process}.
        \shh{

    \noindent{\textbf{Other Experiments.}} We have also performed experiments on the following aspects: (1) time complexity, (2) effect of different pseudo-labelling methods, (3) effect of the number of labeled samples in each domain and (4) effect of overlapping rate of known label spaces between the two domains. For the sake of space, descriptions and results of these experiments are given in the supplementary material.  
    % We empirically compared the complexity of our method with top competitors on Intel E5-2650 CPU, with an Nvidia GTX TITAN X GPU for the  deep method. The time cost of each of the methods on different tasks are shown in Table \ref{table:time}. As shown, the time complexity of CDA is comparable to those for the other \shhh{domain adaptation} methods, while \rev{its accuracy} outperformed \shhh{all} the \shhh{compared} methods.
    
    }

    % \vspace{-0.1cm}
    \section{Conclusion}
    % \vspace{-0.1cm}
      In this paper, we introduced the weakly supervised open-set domain adaptation problem. 
      \rev{This setting extends domain adaptation to the scenarios 
      where 
      an ideal source domain is absent, and we need to let partially labeled domains learn from each other.}
      To address this problem, we propose a novel Collaborative Distribution Alignment approach. 
      Compared to previous works, this study represents the first attempt to enhance overall performance by domain collaboration. 
      Future works could be undertaken to detect the unknown-class samples more effectively
       or to design an adaptive neural network for end-to-end training.
      % Extensive experiments on the \textit{Office} and DukeMTMC-reID datasets showed that CDA achieves state-of-the-art performances overall.

    % \vspace{-0.1cm}
    \section{Acknowledgement}  
    \vspace{-0.1cm}
    This work was supported partially by NSFC (U1811461, 61522115), Guangdong Province Science and Technology Innovation Leading Talents (2016TX03X157).
    % We would like to thank Shisheng Tan and Xuhong An for their encouragement and insightful discussions.
 
    {\small
    \bibliographystyle{ieee}
    \bibliography{cites}
    }
    \end{document}